\documentclass{article}

\usepackage{PRIMEarxiv}

\usepackage[utf8]{inputenc} 
\usepackage[T1]{fontenc}    
\usepackage{hyperref}       
\usepackage{url}            
\usepackage{booktabs}       
\usepackage{amsfonts}       
\usepackage{nicefrac}       
\usepackage{microtype}      
\usepackage{lipsum}
\usepackage{fancyhdr}
\usepackage{caption}
\usepackage{graphicx}       
\graphicspath{{media/}}
\usepackage{amsmath}
\usepackage{multirow}
\usepackage{arydshln}
\usepackage{mdframed}
\usepackage{xcolor}
\usepackage{threeparttable}

\title{XPASS-Vis: A Dataset for Cross-Domain Personalized Image Aesthetic Assessment}

\author{
  Takato Hayashi$^{1}$, Hiroaki Takahara$^{1}$, Candy Olivia Mawalim$^{1}$, \\
  \textbf{Hiromi Narimatsu$^{2}$, Akisato Kimura$^{2}$, Shiro Kumano$^{2}$, Shogo Okada$^{1}$} \\
  \\
  $^{1}$Japan Advanced Institute of Science and Technology, \\
  1-1 Asahidai, Nomi, Ishikawa 923-1292, Japan \\
  $^{2}$Communication Science Laboratories, NTT, Inc., \\
  Atsugi-shi 243-0198, Japan \\
  \texttt{hayashi0884@jaist.ac.jp} \\
}

\begin{document}
\maketitle

\begin{abstract}
Personalized image aesthetic assessment (PIAA) seeks to model, at the individual level, the subjective nature of aesthetic judgments toward artworks and photographs. Aesthetic preference is known to be both deeply personal and partially consistent across visual domains. Yet existing PIAA datasets and methods are largely confined to a single domain, or provide too few samples per annotator within each domain to enable personalization across domains. Consequently, the cross-domain generalization of personalized aesthetic preferences remains largely unexplored. To address this gap, we introduce XPASS-Vis, the first dataset explicitly designed for cross-domain PIAA. XPASS-Vis comprises 6,526 stimuli from three visual domains---art, fashion, and landscape---rated by 129 annotators, yielding 87,836 user--stimulus interactions, each annotated with an overall aesthetic score and nine aesthetic-emotion ratings. Notably, each annotator rated more than 200 stimuli per domain, providing sufficient per-domain coverage to support personalization both within and across domains. Moreover, we establish baseline models for cross-domain PIAA under unsupervised domain adaptation (UDA), where a model trained on a labeled source domain is transferred to an unlabeled target domain. A systematic evaluation of representative UDA approaches shows that the best-performing method recovers approximately 60\% (Spearman's $\rho$ = .28) of the supervised upper bound under a fully unsupervised setting. This provides encouraging evidence that personalized aesthetic preferences are, to a meaningful extent, transferable across visual domains. At the same time, a substantial gap remains, highlighting the need for PIAA-specific adaptation strategies. XPASS-Vis and the accompanying baselines provide a foundation for future research on cross-domain PIAA. All datasets and code will be made publicly available upon acceptance.

\end{abstract}


\begin{figure*}[t]
\centering
\includegraphics[width=1\linewidth]{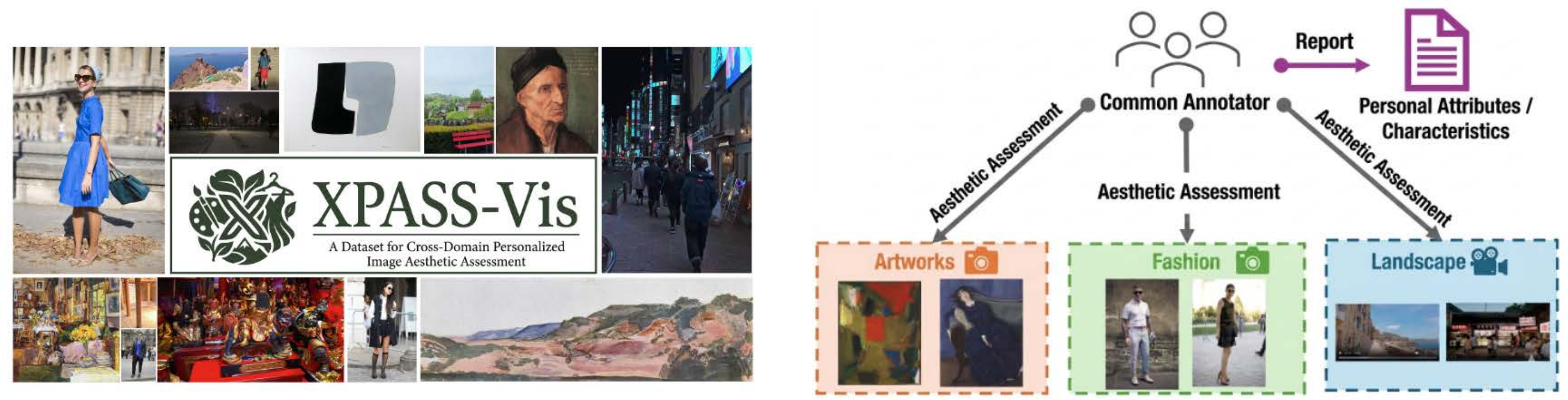}
\caption{Overview of XPASS-Vis. The same 129 annotators evaluate stimuli from three visual domains (artwork, fashion, and landscape) and report their personal attributes (e.g., demographics and personality traits), yielding 87,836 user--stimulus interactions across domains. This shared-annotator design enables cross-domain personalized aesthetic assessment.}
\label{fig:XPASS-Vis}
\end{figure*}

\section{Introduction}
\label{sec:intro}

Understanding and modeling individual aesthetic experiences have become a central goal in \textit{computational aesthetics}. Personalized image aesthetic assessment (PIAA) aims to capture the subjective nature of aesthetic judgments by modeling how individual users perceive and evaluate visual content beyond population-level consensus. PIAA has broad applications in personalized content generation, creative support systems, and recommendation systems.

Aesthetic preference is characterized by two fundamental properties: it is deeply personal, yet consistent across domains. First, aesthetic taste is inherently individual~\cite{yang2022personalizedimageaestheticsassessment, zhang2021lapis, zhu2023personalized, shi2023personalized, wang2026enhancingzeroshotpersonalizedimage}. The same painting can spark admiration in one person and indifference in another. Second, individual aesthetic taste is consistent across domains~\cite{Jacobsen2017-lf, Pham2026_aes}. Each individual holds some preferences that are domain-specific and others that are shared across domains. For instance, a person who values simplicity may be drawn to both minimalist paintings and understated fashion styles---suggesting that such a preference reflects an individual aesthetic value shared across domains rather than a domain-specific one. Yet the extent of this consistency remains untested: existing PIAA datasets are confined to single domains~\cite{zhang2021lapis, Zhong_2025_CVPR} or provide too few per-annotator samples within each domain~\cite{yang2022personalizedimageaestheticsassessment}, leaving current methods unevaluated across domains. It thus remains an open question to what extent machine learning models can transfer an individual's aesthetic preferences across visual domains.

\begin{table*}[!t]
\centering
\caption{Comparison of representative PIAA datasets.}
\label{tab:dataset_comparison}
\setlength{\tabcolsep}{2pt}
\renewcommand{\arraystretch}{1.2}
\scriptsize
\begin{tabular}{lccccccccc}
\toprule
\textbf{Dataset} & \textbf{Domain} & \textbf{Venue} & \textbf{\# Stimuli} & \textbf{\# Annotators} & \textbf{\# Interactions} & \textbf{Per-user} & \textbf{Cross-domain} & \textbf{User attr.} & \textbf{Aest. emotion} \\
\midrule
AADB~\cite{kong2016photoaestheticsrankingnetwork}              & Generic photos  & ECCV'16     & 10{,}000  & 190         & $\sim$50{,}000      & $\sim$263     & $\triangle$ & --         & -- \\
FLICKR-AES~\cite{ren2017personalized}                          & Generic photos  & ICCV'17     & 40{,}000  & 210         & $\sim$200{,}000     & $\sim$952     & $\triangle$ & --         & -- \\
REAL-CUR~\cite{ren2017personalized}                            & Personal photos & ICCV'17     & $\sim$2{,}870 & 14      & $\sim$2{,}870       & $\sim$205     & $\triangle$ & --         & -- \\
PARA~\cite{yang2022personalizedimageaestheticsassessment}      & Generic photos  & CVPR'22     & 31{,}220  & 438         & $\sim$807{,}000     & $\sim$1{,}843 & $\triangle$ & \checkmark & $\triangle$ \\
LAPIS~\cite{zhang2021lapis}                                    & Art             & CVPR--WS'25 & 11{,}723  & 547         & $\sim$281{,}352     & $\sim$514     & --          & \checkmark & -- \\
\midrule
\textbf{XPASS-Vis} & \textbf{Art, Fashion, Landscape} & \textbf{Ours} & \textbf{6{,}526} & \textbf{129} & \textbf{87{,}836} & \textbf{$\sim$681} & \checkmark & \checkmark & \checkmark \\
\bottomrule
\end{tabular}
\vspace{2pt}
\begin{minipage}{\linewidth}
\footnotesize
\textit{Interactions}: number of user--stimulus rating pairs.
\textit{Per-user}: average number of stimuli rated per annotator.
\checkmark: supported;\quad --: not supported;\quad $\triangle$: partially supported.
For \textit{Cross-domain}, $\triangle$ denotes that images span multiple domains but without sufficient per-user samples per domain for cross-domain analysis.
For \textit{Aest. emotion}, $\triangle$ denotes that only a single dominant basic-emotion is provided per image, rather than a multi-item aesthetic-emotion.
\end{minipage}
\end{table*}

To address the lack of data infrastructure for studying cross-domain PIAA, we present \textbf{XPASS-Vis} (Fig.~\ref{fig:XPASS-Vis}), the first dataset explicitly designed for PIAA across visual domains. XPASS-Vis consists of 6,526 stimuli drawn from three domains: artwork, fashion, and landscape. We choose these domains because they represent major categories of everyday aesthetic experience---spanning culturally constructed to naturally occurring stimuli---while plausibly sharing aesthetic preferences that transfer across them. A total of 129 annotators participated in the study. For each stimulus, annotators provided ten ratings that capture two complementary facets of the aesthetic experience: a single \emph{overall aesthetic assessment}, a holistic judgment of how aesthetically pleasing the stimulus is, and nine \emph{aesthetic-emotion} items, each capturing a distinct emotion the viewer actually felt in response to the stimulus (e.g., nostalgia). In total, XPASS-Vis contains 87,836 user--stimulus interactions.

XPASS-Vis enables a variety of new research tasks in cross-domain PIAA. In this work, we focus on domain adaptation, in which a model trained on a source domain is adapted to target domains by aligning domain-specific feature distributions while preserving shared aesthetic preferences. This task is of practical importance because collecting sufficient labeled data in every domain is prohibitively expensive, and a cold-start problem arises in new domains where no user rating history is available. We therefore evaluate domain adaptation under unsupervised scenarios, where no labeled target-domain data is available, by systematically applying a range of representative techniques as baseline models. Our aim is to identify which adaptation strategies are most effective for transferring aesthetic preferences across visual domains.

Our contributions are twofold:
\begin{itemize}
\item We release XPASS-Vis, the first dataset for cross-domain PIAA, which will be publicly available for research purposes upon acceptance.
\item We establish a new benchmark task for cross-domain PIAA, and systematically evaluate a range of domain adaptation techniques as baseline models, providing insights into the challenges of transferring aesthetic preferences across visual domains.
\end{itemize}

\begin{table*}[t]
\centering
\caption{Representative UDA methods evaluated as baselines for cross-domain PIAA.}
\label{tab:uda_methods}
\small
\begin{tabular}{llccp{0.55\linewidth}}
\toprule
Method & Venue & Category & Reg. & Brief description \\
\midrule
DANN~\cite{Ganin2016} & JMLR'16 & Adv. & -- & Learns domain-invariant features via a gradient-reversal layer and a domain discriminator trained adversarially against the feature extractor. \\
CDAN~\cite{Long2017-tu} & NeurIPS'18 & Adv. & -- & Conditions the adversarial alignment on the multilinear outer product of features and classifier predictions to align joint distributions. \\
ALDA~\cite{chen2020adversariallearnedlossdomainadaptation} & AAAI'20 & Adv. & -- & Derives a corrected loss from the domain discriminator's output to mitigate pseudo-label noise during adversarial alignment. \\
DeepJDOT~\cite{damodaran2018deepjdotdeepjointdistribution} & ECCV'18 & Disc.\ (OT) & -- & Aligns the joint distribution of features and labels by minimizing an optimal-transport distance between source and target. \\
JUMBOT~\cite{Fatras2021-qw} & ICML'21 & Disc.\ (OT) & -- & Extends JDOT with unbalanced mini-batch optimal transport, providing robustness to outlier samples and class imbalance. \\
DeepCORAL~\cite{Sun8_35} & ECCV-WS'16 & Disc.\ (FA) & -- & Aligns second-order feature statistics by minimizing the distance between source and target covariance matrices. \\
RSD~\cite{Chen2021-ua} & ICML'21 & Disc.\ (FA) & \checkmark & Minimizes the principal-angle distance between source and target feature subspaces; designed specifically for regression. \\
DARE-GRAM~\cite{nejjar2023domain} & CVPR'23 & Disc.\ (FA) & \checkmark & Aligns the scale and angle of inverse Gram matrices, yielding regression-aware feature alignment. \\
\bottomrule
\end{tabular}
\\[2pt]
\raggedright\footnotesize Adv.: adversarial-learning based; Disc.: discrepancy-based (OT: optimal transport, FA: feature alignment); Reg.: regression-aware design.
\end{table*}

\section{Related Works}
\label{sec: related}
We briefly review two lines of research most relevant to this work: personalized image aesthetic assessment (PIAA) and unsupervised domain adaptation (UDA).

\subsection{Personalized Image Aesthetic Assessment}
While general image aesthetic assessment (GIAA) predicts population-aggregated scores across annotators~\cite{Talebi2018-tj, chen2025role, behrad2025charmmissingpiecevit, yang2026finegrainedigiaa} and benefits from large-scale benchmarks~\cite{murray2012ava, He_ijcai2022p132, yi2023artisticimageaestheticsassessment, nieto2022understandingaestheticslanguagephoto}, aesthetic judgments are inherently subjective and vary considerably across individuals~\cite{yang2022personalizedimageaestheticsassessment, zhang2021lapis}. Applying GIAA models directly to individual users therefore yields poor personalization~\cite{yang2022personalizedimageaestheticsassessment}, as population-level predictions fail to capture the idiosyncratic preferences of each person. PIAA addresses this limitation by modeling aesthetic judgment at the individual level. Recent methods incorporate user attributes—such as demographics and personality traits—together with image features to predict user-specific preferences~\cite{zhu2023personalized, shi2023personalized, wang2026enhancingzeroshotpersonalizedimage}. However, only a few datasets have collected user attributes, and the many existing PIAA studies~\cite{zhu2023personalized, shi2023personalized, wang2026enhancingzeroshotpersonalizedimage, chen2025role} therefore rely on either PARA~\cite{yang2022personalizedimageaestheticsassessment} or LAPIS~\cite{zhang2021lapis} as their primary benchmark. Empirical evidence further suggests that aesthetic appreciation has both domain-specific and domain-general components~\cite{Jacobsen2017-lf, Pham2026_aes}, implying that individual taste should partly transfer across visual domains. However, existing PIAA methods and datasets have so far been developed within a single domain, or do not provide sufficient per-user samples within each domain, making them unsuitable for cross-domain PIAA. Although smaller in scale than prior single-domain datasets, XPASS-Vis is, to the best of our knowledge, the first practical dataset that supports cross-domain PIAA by providing sufficient per-user samples within each domain ($>200$) across three domains.

\begin{table*}[t]
\centering
\caption{Overview of demographics and collected attributes for the 129 annotators.}
\vspace{6pt}
\label{tab:annotator_attributes}
\small
\begin{tabular}{p{2.5cm} p{8cm} p{5cm}}
\hline
\textbf{Attribute} & \textbf{Details / Scale} & \textbf{Summary (N=129)} \\
\hline
Age 
& Numeric: Integer value entered directly by participants. 
& Mean: 28.7 $\pm9.5$, Range: 20--67 \\
\hline
Gender 
& Selection: Male, Female, Other. 
& Male: 69(53.5\%), Female: 59(45.7\%), Other: 1(0.8\%) \\
\hline
Nationality 
& Selection: All countries and regions selectable. 
& Japan: 87(67.4\%), China: 41(31.8\%), Korea: 1(0.8\%) \\
\hline
Academic History 
& Selection: Junior High School, High School(HS), Vocational School(VS), Junior College(JC), Technical College(TC), Bachelor(BA), Graduate School(GS), Other. 
& BA: 68(52.7\%), GS: 41(31.8\%), HS: 9(7.0\%), TC: 5(3.9\%), JC: 5(3.9\%), VS: 1(0.8\%) \\
\hline
Personality Traits 
& Likert scale: Ten-Item Personality Inventory (TIPI); 1 = strongly disagree, 7 = strongly agree. 
& See Fig.~\ref{fig:user_stats} \\
\hline
Domain Interest 
& Likert scale: Self-reported interest level for each domain; 1 = not interested at all, 7 = extremely interested.
& See Fig.~\ref{fig:user_stats} \\
\hline
Domain Education
& Free text: Participants reported educational experience related to Art, Fashion, or Photo/Video, including the type of institution and the duration (years/months) attended. 
& Number of annotators with formal education: Art:8(6.2\%), Fashion:2(1.6\%), Photo/Video:1(0.8\%) \\
\hline
\end{tabular}
\end{table*}

\subsection{Unsupervised Domain Adaptation}
UDA aims to transfer a model trained on a labeled source domain to an unlabeled target domain by mitigating distributional discrepancy between the two. Existing approaches can be broadly grouped into two categories (Table~\ref{tab:uda_methods}): \emph{adversarial-learning-based} methods and \emph{discrepancy-based} methods. Adversarial-learning-based methods aim to learn domain-invariant representations via adversarial training, with typical methods including DANN~\cite{Ganin2016}, CDAN~\cite{Long2017-tu}, and ALDA~\cite{chen2020adversariallearnedlossdomainadaptation}. Discrepancy-based methods explicitly minimize a domain distance using discrepancy metrics such as optimal transport distance or feature correlation alignment. Representative methods based on optimal transport include DeepJDOT~\cite{damodaran2018deepjdotdeepjointdistribution} and JUMBOT~\cite{Fatras2021-qw}, while feature alignment methods include DeepCORAL~\cite{Sun8_35}, with regression-oriented variants such as RSD~\cite{Chen2021-ua} and DARE-GRAM~\cite{nejjar2023domain}. Beyond these, source-free UDA has also been studied, where only a pretrained source model—not the source data itself—is available during adaptation. In this work,  we focus on the more standard and higher-performing setting in which both labeled source data and unlabeled target data are accessible, and adopt representative methods as baselines for cross-domain PIAA.

\section{Data Collection}
\label{sec:data}
This section describes how XPASS-Vis was constructed. We first introduce the participants and the questionnaires used to collect their personal attributes (Sec.~\ref{sec:participants}), then describe how visual stimuli were selected from three domains---artwork, fashion, and landscape (Sec.~\ref{sec:stim}). We then detail the annotation procedure, including the items rated for each stimulus and the protocol used to ensure consistent aesthetic judgment across domains (Sec.~\ref{sec:ann}). Finally, we describe the quality control pipeline applied to filter unreliable annotators and low-quality responses, and report basic statistics of the resulting dataset (Sec.~\ref{sec:qc} and~\ref{sec:analysis}).

\subsection{Participants}
\label{sec:participants}
A total of 145 annotators participated in the study. Of these, 125 were recruited through the mailing list of the authors' institution; as nearly 70\% were male, we additionally recruited 20 female participants through a staffing agency. Participation eligibility required sufficient Japanese proficiency, as all tasks were administered in Japanese. Before annotation, each participant completed a questionnaire collecting demographic information, personality traits (TIPI~\cite{Gosling2003-ct, Oshio2012-bf}), and domain-related attributes including interest and education in art, fashion, and photo/video. We note that, for the landscape domain, the interest and education attributes were measured using the corresponding photo/video items, as the questionnaire included no items specific to landscape photography. These proxy measures are less direct than those for art and fashion, so analyses and discussions involving this attribute should be interpreted with caution. Compensation followed institutional wage standards for internal participants and platform policies for external participants. All participants provided written informed consent, and the study was approved by the Research Ethics Committee of the authors' institution.

\subsection{Stimuli Selection}
\label{sec:stim}
XPASS-Vis consists of three domains: artwork, fashion, and landscape, representing major categories of everyday aesthetic experience with distinct visual characteristics and functional roles. Together, these domains span a broad range of aesthetic experience, from culturally constructed to naturally occurring, and from contemplative to desire-driven. At the same time, meaningful cross-domain relationships can be expected among them. For the art domain, we included all stimuli from the test split of the LAPIS dataset~\cite{zhang2021lapis}, yielding 2,345 artworks spanning 26 artistic styles (from Renaissance to Minimalism) and 7 genres (e.g., abstract). For the fashion domain, we used the Clothing Co-Parsing (CCP) dataset~\cite{yang2014clothing}; after excluding images with two or more central figures, 2,082 high-resolution fashion photographs with diverse clothing styles remained. For the landscape domain, we randomly sampled 2,099 one-minute clips spanning 101 countries from Sekai-Real-Walking-HQ~\cite{li2025sekai}, a large-scale collection of over 18,000 walking video clips. Although the latter two datasets were not originally designed for aesthetic assessment, both provide high visual quality and diverse content suitable for this purpose.

\begin{figure*}[t]
    \centering
    \includegraphics[width=1\linewidth]{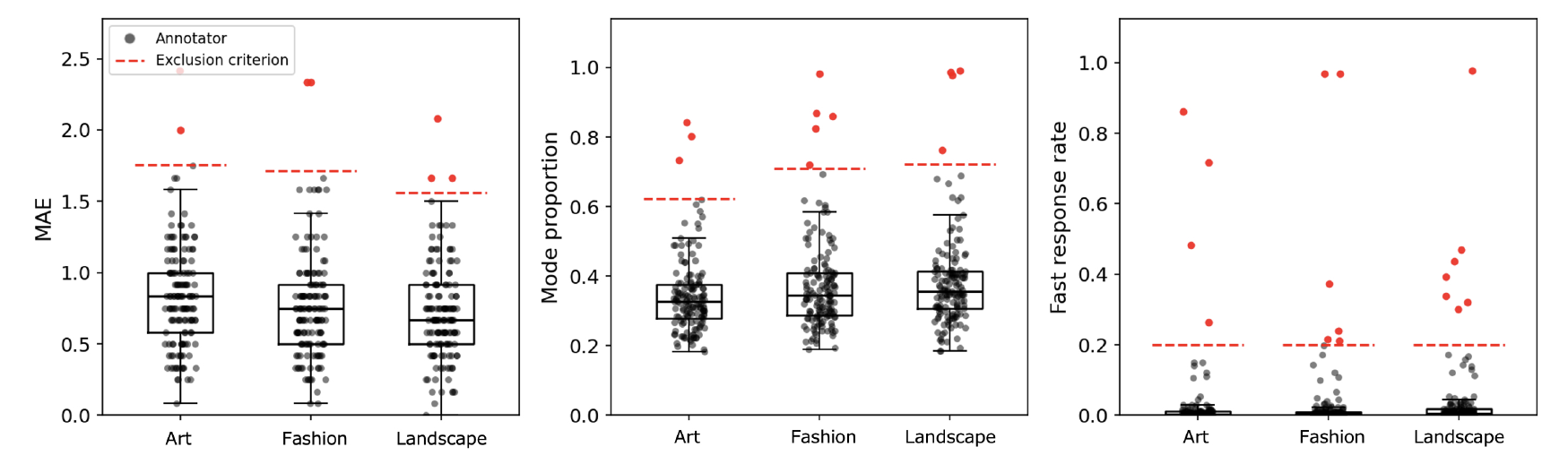}
    \vspace{0.2em}
\caption{%
    Quality control metrics per annotator across three stimulus domains (art, fashion, landscape).
    Each dot represents one annotator; red dashed lines indicate the exclusion criterion
    ($\mu + 2.5\,\sigma$ for $p_{\text{mode}}$ and MAE; $r_{\text{fast}} > 0.20$).
    Annotators exceeding the criterion in any domain were excluded.%
}
    \label{fig:annotator_filtering}
\end{figure*}

\begin{figure}[t]
\centering
\begin{minipage}[t]{0.48\linewidth}
    \centering
    \includegraphics[width=\linewidth]{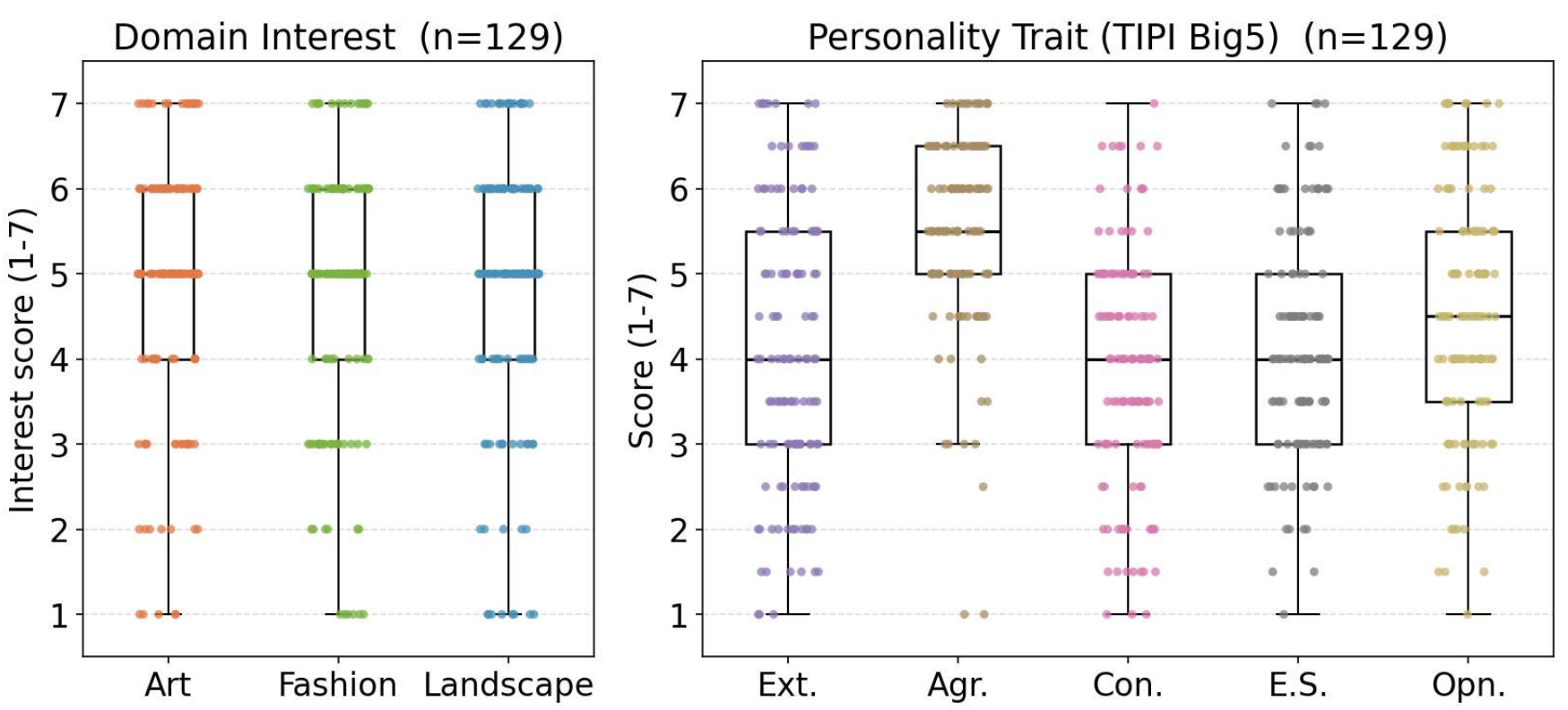}
    \caption{%
        Distributions of self-reported domain interest (left) and TIPI BigFive personality traits (right) across N = 129 participants, on a common 1--7 Likert scale. BigFive abbreviations: Ext. = Extraversion, Agr. = Agreeableness, Con. = Conscientiousness, E.S. = Emotional Stability, Opn. = Openness.
    }
    \label{fig:user_stats}
\end{minipage}
\hfill
\begin{minipage}[t]{0.44\linewidth}
    \centering
    \includegraphics[width=\linewidth]{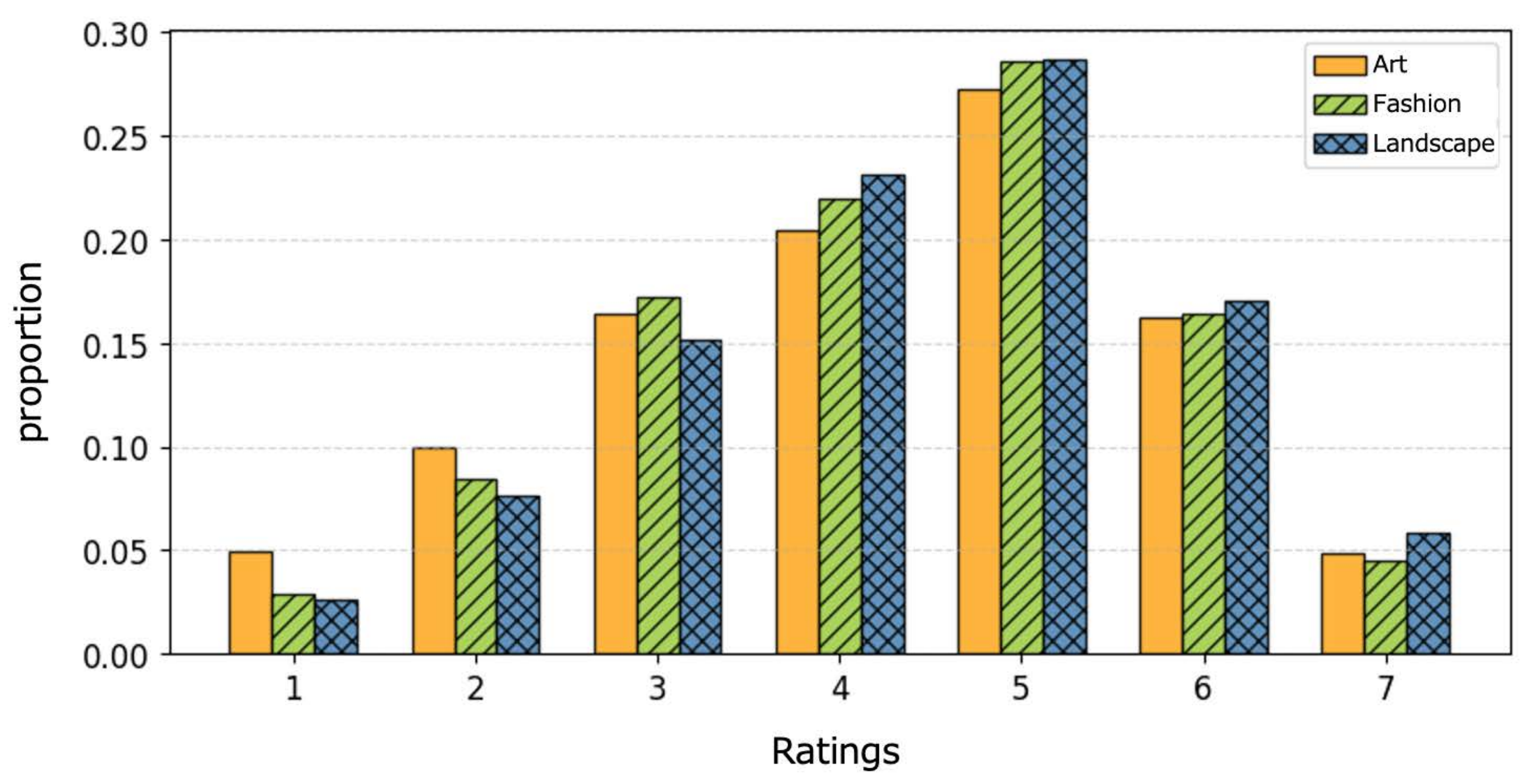}
    \caption{Distribution of overall aesthetic assessment across three domains.}
    \label{fig:aesthetic_hist}
\end{minipage}
\end{figure}

\subsection{Annotation Procedure}
\label{sec:ann}
We constructed ten annotation sets, each containing approximately 230 artworks, 210 fashion images, and 210 video clips, and randomly assigned one set to each annotator. Within each domain, 12 stimuli were randomly selected and presented twice to assess test–retest reliability. All annotations were conducted through a custom-built web-based platform (Fig.~\ref{fig:annotation_tool}), allowing remote completion. The annotation process required approximately 15 hours for roughly 650 stimuli, with annotators given two weeks to proceed at their own pace. To promote consistent aesthetic criteria across domains, the stimuli in each domain were divided into subsets of approximately 30, which were then interleaved across domains rather than presented one domain at a time.

For each stimulus, annotators responded to ten items: one overall aesthetic assessment and nine aesthetic emotion items selected from AESTHEMOS~\cite{Schindler2017-kz}, a validated questionnaire covering diverse aesthetic emotions. We selected the item with the highest factor loading from each of the seven higher-order factors, plus \textit{like} and \textit{beautiful} as canonical markers of aesthetic judgment, yielding: \textit{'Liked it'}, \textit{'I found it beautiful'}, \textit{'I found it distasteful'}, \textit{'Was impressed'}, \textit{'Challenged me intellectually'}, \textit{'Motivated me to act'}, \textit{'Made me feel nostalgic'}, \textit{'Made me sad'}, and \textit{'Amused me'}. The items were translated into Japanese by the authors through discussion, taking care to faithfully preserve the meaning of the original scale. Following AESTHEMOS, the instructions emphasized that annotators should base their ratings on the emotions they personally felt, rather than on the emotions that the stimulus was attempting to convey. Each aesthetic emotion was rated on a 5-point Likert scale (1 = \textit{Not at all}, 5 = \textit{Very strongly})~\cite{Schindler2017-kz}, while the overall aesthetic assessment used a 7-point scale (1 = \textit{Highly unaesthetic}, 7 = \textit{Highly aesthetic}) following LAPIS~\cite{zhang2021lapis}. The aesthetic emotion items were included to enable fine-grained analysis of how overall assessments relate to specific aesthetic emotions in future research; the present study focuses solely on the overall aesthetic assessment.

\begin{table}[t]
\centering
\caption{The composition of the XPASS-Vis dataset after quality control.}
\label{tab:dataset_stats}
\begin{tabular}{lrrcccc}
\toprule
Domain    & Stimulus & Interactions & ICC(1,1) & $r$ & MAE \\
\midrule
Art       & 2,345    & 31,543       & 0.24     & 0.57        & 0.80 \\
Fashion   & 2,082    & 28,154       & 0.19     & 0.58        & 0.75 \\
Landscape & 2,099    & 28,139       & 0.23     & 0.59        & 0.71 \\
\midrule
Total     & 6,526    & 87,836       & —        & —           & — \\
\bottomrule
\end{tabular}
\begin{minipage}{\linewidth}
\vspace{4pt}
\footnotesize
ICC(1,1): inter-annotator agreement, measuring the consistency of aesthetic ratings across annotators; Pearson $r$: intra-annotator agreement (test--retest reliability), measuring the rank-order correlation between a annotator's first and second ratings of the same stimuli; MAE: intra-annotator agreement (test--retest reliability), measuring the mean absolute error between first and second ratings of the same stimuli.
\end{minipage}
\end{table}

\subsection{Quality Control}
\label{sec:qc}
We applied a quality control pipeline to filter unreliable annotators and low-quality responses (see Fig.~\ref{fig:annotator_filtering}). After applying all criteria, the dataset was reduced from 145 annotators and 99{,}901 interactions to 129 annotators and 87{,}836 interactions.

\textbf{Annotator-Level Filtering.} Annotator reliability was assessed across domains using three criteria. First, we computed the mean absolute error (MAE) between repeated ratings of the same stimuli as a measure of test--retest reliability; annotators whose MAE exceeded the group mean $+ 2.5 \times \mathrm{SD}$ in any domain were flagged. The $+ 2.5 \times \mathrm{SD}$ criterion is a widely adopted convention for outlier removal~\cite{Yang2019-ji}. Second, we computed the mode proportion ($p_{\mathrm{mode}}$), defined as the fraction of ratings equal to the most frequent value within each domain; annotators whose $p_{\mathrm{mode}}$ exceeded the group mean $+ 2.5 \times \mathrm{SD}$ in any domain were flagged as providing near-constant ratings. Third, we computed the fast response rate ($r_{\mathrm{fast}}$), defined as the proportion of responses falling below domain-specific thresholds (10 seconds for still images; 30 seconds for video clips), determined empirically by the authors performing the tasks themselves; annotators with $r_{\mathrm{fast}} > 0.20$ in any domain were flagged, as exceeding this threshold would leave insufficient stimuli to construct reliable training and test sets. A total of 16 annotators were excluded.

\textbf{Sample-Level Filtering.} Individual responses below the response time thresholds were removed regardless of annotator status, resulting in 3{,}254 excluded responses.

\subsection{Data Analysis}
\label{sec:analysis}
Table~\ref{tab:dataset_stats} summarizes the composition of the final XPASS-Vis dataset after quality control, while Table~\ref{tab:annotator_attributes} presents the demographic and personality attributes of the 129 annotators.

\textbf{Annotator Demographics.} As shown in Table~\ref{tab:annotator_attributes}, annotators ranged in age from 20 to 67 years, with a near-balanced gender distribution. The majority held at least a university degree (84.5\%), and nationality was predominantly Japanese (67.4\%) and Chinese (31.8\%), reflecting the nature of the authors' institution. Very few annotators reported formal education in the evaluated domains, indicating that the dataset represents non-expert, general audiences. Fig.~\ref{fig:user_stats} shows the distributions of personality traits and domain interest levels. Interest was consistently high across all domains (median $\approx 5$), indicating that annotators had substantial interest in art, fashion, and photo/video.

\textbf{Overall Aesthetic Assessment and Agreement.} Fig.~\ref{fig:aesthetic_hist} shows the distribution of aesthetic scores. Scores spanned the full range of the 7-point scale, and mean scores were comparable across domains (art: 3.23, fashion: 3.32, landscape: 3.42). Inter-annotator agreement (ICC(1,1)) was low across all domains (art: 0.24; fashion: 0.19; landscape: 0.23), reflecting the inherently subjective nature of aesthetic assessment. Intra-annotator agreement was assessed using both Pearson $r$ and MAE between repeated ratings of the same stimuli. Pearson $r$ (art: 0.57; fashion: 0.58; landscape: 0.59) indicates relatively consistent individual rank-ordering over time. MAE (art: 0.80; fashion: 0.75; landscape: 0.71) indicates that individual annotators assigned similar absolute scores on repeated viewings. Statistics for all items, including the aesthetic-emotion items, are provided in Table~\ref{tab:item_stats} and Fig.~\ref{fig:rating_histograms}.

\section{Experiment}
This section describes the experimental setup used to benchmark within-domain and cross-domain PIAA on XPASS-Vis. We first detail the preprocessing applied to the visual stimuli and the encoding of user and image attributes (Sec.~\ref{sec:exp-preprocessing}). We then introduce the training pipeline, covering the GIAA model, the two-phase personalized models, and the zero-shot LLM baselines (Sec.~\ref{sec:exp-pipeline}), followed by the unsupervised domain adaptation methods evaluated for cross-domain transfer (Sec.~\ref{sec:exp-uda}). Finally, we present the evaluation procedure and metrics (Sec.~\ref{sec:exp-eval}) and the training setup and implementation details (Sec.~\ref{sec:exp-training}).\\

\begin{figure*}[t]
\centering
\begin{minipage}[t]{0.53\textwidth}
\centering
\captionof{table}{Quantitative image properties (QIPs) feature list}
\label{tab:qip_features}
\tiny
\vspace{3pt}
\begin{tabular}{clp{3.5cm}}
\toprule
\textbf{\#} & \textbf{Feature} & \textbf{Description} \\
\midrule
\multicolumn{3}{l}{\textit{Basic Image Properties (6-dim)}} \\
\midrule
1  & Image size         & Total number of pixels \\
2  & Aspect ratio       & Width-to-height ratio \\
3  & RMS contrast       & Root mean square contrast \\
4  & Luminance entropy  & Entropy of luminance distribution \\
5  & Complexity         & Overall image complexity \\
6  & Edge density       & Density of edge pixels \\
\midrule
\multicolumn{3}{l}{\textit{Color Properties (19-dim)}} \\
\midrule
7      & Color entropy     & Entropy of color distribution \\
8--10  & RGB mean          & Mean values of R/G/B channels \\
11--13 & Lab mean          & Mean values of L/a/b channels \\
14--16 & HSV mean          & Mean values of H/S/V channels \\
17--19 & RGB std           & Standard deviation of R/G/B channels \\
20--22 & Lab std           & Standard deviation of L/a/b channels \\
23--25 & HSV std           & Standard deviation of H/S/V channels \\
\midrule
\multicolumn{3}{l}{\textit{Composition and Balance (5-dim)}} \\
\midrule
26     & Mirror symmetry   & Degree of mirror symmetry \\
27     & DCM distance      & Distance in density-corrected model \\
28--29 & DCM position      & x/y position in DCM \\
30     & Balance           & Overall compositional balance \\
\midrule
\multicolumn{3}{l}{\textit{Symmetry (3-dim)}} \\
\midrule
31 & CNN symmetry (LR)    & CNN-based left--right symmetry \\
32 & CNN symmetry (TB)    & CNN-based top--bottom symmetry \\
33 & CNN symmetry (LR+TB) & Combined CNN symmetry \\
\midrule
\multicolumn{3}{l}{\textit{Texture and Frequency Properties (8-dim)}} \\
\midrule
34 & Fourier slope       & Slope of Fourier spectrum \\
35 & Fourier sigma       & Sigma parameter of Fourier analysis \\
36 & 2D fractal dim.     & Two-dim fractal dimension \\
37 & 3D fractal dim.     & Three-dim fractal dimension \\
38 & Self-similarity (PHOG) & Self-similarity via PHOG \\
39 & Self-similarity (CNN)  & CNN-based self-similarity \\
40 & Anisotropy          & Measure of directional dependence \\
41 & Homogeneity         & Measure of spatial uniformity \\
\midrule
\multicolumn{3}{l}{\textit{Visual Complexity (4-dim)}} \\
\midrule
42 & EOE (1st order)    & First-order edge orientation entropy \\
43 & EOE (2nd order)    & Second-order edge orientation entropy \\
44 & Sparsity           & Sparsity of visual features \\
45 & Variability        & Temporal/spatial variability \\
\bottomrule
\end{tabular}
\end{minipage}\hfill
\begin{minipage}[t]{0.44\textwidth}
\centering
\vspace{0pt}
\includegraphics[width=\linewidth]{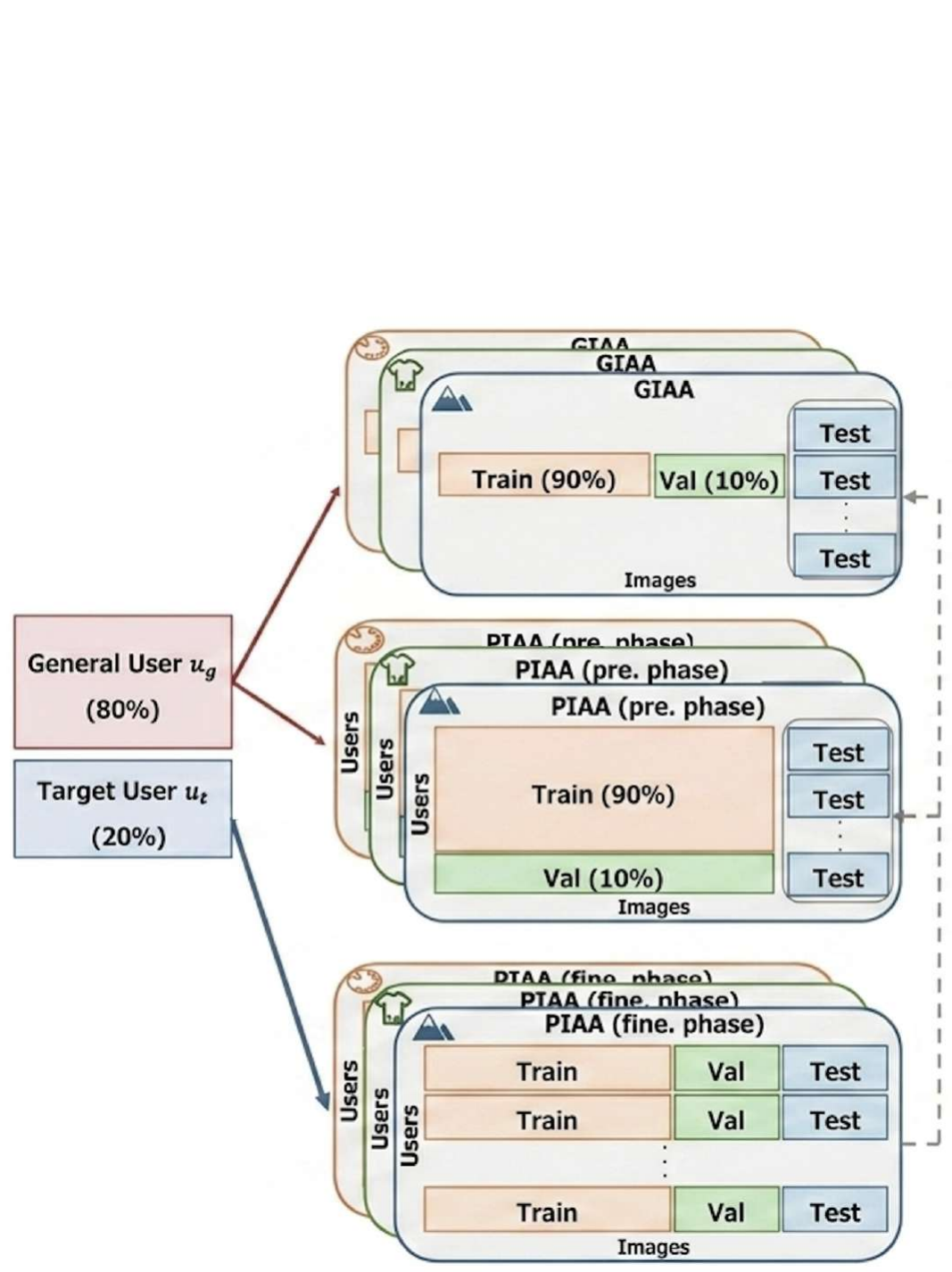}
\captionof{figure}{Data splitting scheme. Annotators are split into general users $u_g$ (80\%) and target users $u_t$ (20\%) via 5-fold cross-validation. GIAA uses image-level splits of $u_g$'s aggregated ratings; PIAA pre.\ phase uses a user-level split of $u_g$; PIAA fine.\ phase fine-tunes per target user (100 train samples). All stages share the same $u_t$ test set (dashed arrows), applied independently to each of the three domains.}
\label{fig:data_split}
\end{minipage}
\end{figure*}

\subsection{Preprocessing}
\label{sec:exp-preprocessing}
\textbf{Video-to-Image Conversion.}
The landscape domain consists of video clips. Because these clips are one-minute recordings of walking, they exhibit limited visual variation across frames, so a single frame can largely represent an entire clip. Given this, together with the finding that image-based models outperformed video-based ones (see I3D in Table~\ref{tab:comp_backbone}), we randomly sampled a single frame from each clip to ensure modality consistency across domains and to reduce computational cost.

\textbf{User attributes.}
Following prior work~\cite{zhang2021lapis}, annotator attributes are encoded as one-hot vectors and concatenated into a fixed-dimensional feature vector. \textit{Personality traits} (TIPI; Q1--Q10) and \textit{domain interest scores} are each encoded as 7-dim one-hot vectors, yielding 70- and 21-dim vectors, respectively. \textit{Age} is discretized into five age bins (20--28, 29--37, 38--47, 48--56, and 57--67) and encoded as a 5-dim one-hot vector. \textit{Gender} (3-dim), \textit{academic history} (6-dim), and \textit{nationality} (3-dim) are encoded via standard one-hot encoding, and \textit{domain education} as three binary vectors of dimension 2. The concatenation yields a 114-dim feature vector per user.

\textbf{Image attributes.}
Following prior work~\cite{zhang2021lapis}, we extract 45 low-level aesthetic features (Table~\ref{tab:qip_features}) per image using a quantitative image properties (QIPs) toolbox~\cite{Redies2025-wq}, covering \textit{basic image properties} (6-dim), \textit{color statistics} (19-dim), \textit{composition and balance} (5-dim), \textit{symmetry} (3-dim), \textit{texture and frequency properties} (8-dim), and \textit{visual complexity} (4-dim).

\subsection{Training Pipeline}
\label{sec:exp-pipeline}
\subsubsection{General Image Aesthetic Assessment (GIAA).}
GIAA aims to predict aesthetic assessment shared across users by learning the distribution of ratings aggregated from multiple annotators. We adopt \textbf{NIMA}~\cite{Talebi2018-tj} as our GIAA model trained with Earth Mover's Distance (EMD) loss. The backbone is CLIP (ViT-B/16)~\cite{radford2021learningtransferablevisualmodels}, kept frozen during training. This backbone consistently achieves the best performance across all domains (see Table~\ref{tab:comp_backbone}). NIMA is trained to map an input image $I$ to an aesthetic score distribution over the 7-point scale:
\begin{equation}
    \mathbf{p}^{\text{GIAA}} = \text{NIMA}(I).
\end{equation}
At inference, the expected value of the distribution is used as the predicted score.

\subsubsection{Personalized Image Aesthetic Assessment (PIAA).}
Building on the pretrained GIAA model, we introduce a two-phase training pipeline that progressively shifts the model from population-level modeling to individual-level modeling. Given an image $I$, user attributes $\mathbf{a}_u$, and image attributes $\mathbf{q}_I$, a PIAA model predicts a personalized scalar score for user $u$:
\begin{equation}
    \hat{s}^{\text{PIAA}}_{u} = f_{\psi}(I, \mathbf{q}_I, \mathbf{a}_u).
\end{equation}
We evaluate two representative PIAA models: \textbf{MIR}~\cite{zhu2023personalized}, which constructs multi-attribute interactions between subjective user attributes and objective image attributes, and \textbf{ICI}~\cite{shi2023personalized}, which models user--image attribute interactions via graph neural networks. Note that in their original formulations, both MIR and ICI use only hand-crafted QIP features $\mathbf{q}_I$ as the visual input to the user--image attribute interaction module; we introduce a minor modification that additionally feeds the learned image embedding extracted from the GIAA backbone into the interaction module, so that both low-level QIP features and high-level semantic features contribute to personalization. Both models are trained with Mean Squared Error (MSE) loss. Target users are held out from training at the data-splitting stage and are never used except during Phase 2. Specifically, in Phase 1 (population-level pretraining; \textit{pre.}), the personalization module $f_\psi$ is introduced on top of the GIAA model and trained on data from multiple general users, learning population-level patterns of user-dependent preference without accessing any target-user data. In Phase 2 (personalized fine-tuning; \textit{fine.}), the pretrained model is fine-tuned for each target user using a small number of that user's ratings (100 samples), which calibrates the model to individual taste while retaining the shared aesthetic representations acquired during Phase 1.

\subsubsection{Proprietary LLMs.}
We further evaluate proprietary LLMs on both GIAA and PIAA tasks via zero-shot inference. Specifically, we evaluate \textbf{GPT-5.4}, \textbf{Claude Opus 4.6}, and \textbf{Gemini 3.0 Flash}. For GIAA, given an image, the model is prompted to predict the aesthetic score distribution aggregated across multiple annotators. For PIAA, given an image together with a target user's personal attributes, the model is prompted to predict the score that the user would assign. The full prompt template is provided in Fig.~\ref{fig:prompt_PIAA}.

\subsection{Domain Adaptation Method}
\label{sec:exp-uda}
We evaluate cross-domain generalization under an unsupervised domain adaptation (UDA) setting, where the model is trained on a labeled source domain and transferred to an unlabeled target domain. Concretely, at every training stage---GIAA pretraining as well as the PIAA pre.\ and fine.\ phases---the model jointly accesses \emph{labeled} samples from the source domain and \emph{unlabeled} samples from the target domain, and no target-domain label is ever used; the supervised regression loss is computed solely on source-domain labels, while the target domain contributes only through the (label-free) domain-alignment objective. 

We evaluate the following domain adaptation methods: DANN~\cite{Ganin2016}, CDAN~\cite{Long2017-tu}, and ALDA~\cite{chen2020adversariallearnedlossdomainadaptation} as adversarial methods; and DeepJDOT~\cite{damodaran2018deepjdotdeepjointdistribution}, JUMBOT~\cite{Fatras2021-qw}, DeepCORAL~\cite{Sun8_35}, RSD~\cite{Chen2021-ua}, and DARE-GRAM~\cite{nejjar2023domain} as discrepancy-based methods. Unless otherwise noted, each method is applied to GIAA and both PIAA phases, in each case using labeled source-domain samples together with unlabeled target-domain samples. Note that RSD and DARE-GRAM are designed specifically for regression and are therefore applied only to the PIAA phases; for the GIAA model, we instead use DeepCORAL, the best-performing method among those applicable to GIAA. We refer the reader to Table~\ref{tab:uda_methods} for details on each method, and to Sec.~\ref{sec:uda_hparam} for implementation details specific to adapting each method to our task, including hyperparameter configurations.

\subsection{Evaluation Procedure}
\label{sec:exp-eval}
We employ 5-fold cross-validation at the \emph{user level}. In each fold, 20\% of annotators form the target user pool $\mathcal{U}_{\mathrm{t}}$ and the remainder form the general user pool $\mathcal{U}_{\mathrm{g}}$; all reported results are based on predictions for $\mathcal{U}_{\mathrm{t}}$. An overview is illustrated in Fig.~\ref{fig:data_split}. In \textbf{GIAA}, ratings from $\mathcal{U}_{\mathrm{g}}$ are aggregated into per-image score histograms, and images are split at the \emph{image level} into training and validation subsets (90\%/10\%). In \textbf{PIAA}, $\mathcal{U}_{\mathrm{g}}$ is further split at the \emph{user level} into $\mathcal{U}_{\mathrm{g}}^{\mathrm{train}}$ (90\%) and $\mathcal{U}_{\mathrm{g}}^{\mathrm{val}}$ (10\%). Non-aggregated user--image pairs from $\mathcal{U}_{\mathrm{g}}^{\mathrm{train}}$ train the interaction module in the \textit{pre.} phase, while those from $\mathcal{U}_{\mathrm{g}}^{\mathrm{val}}$ are used for early stopping, reflecting transfer to unseen users. In the subsequent \textit{fine.} phase, each target user's ratings are split into 100 for training, 50 for testing, and the remainder (approximately 50) for validation.

Performance is evaluated using SCC (Spearman's rank correlation coefficient) and CCC (Concordance Correlation Coefficient), computed per user and averaged across all 129 annotators. SCC, a standard metric in PIAA research, measures the rank-order correlation between predicted and ground-truth scores. We additionally report CCC to complement SCC: beyond rank correlation, it also captures the absolute agreement between predicted and ground-truth scores, making it sensitive not only to the ordering of stimuli but also to annotator-specific rating tendencies such as scale usage or response style.

\subsection{Training Setup}
\label{sec:exp-training}
All models use the AdamW optimizer with a learning rate of $1 \times 10^{-5}$. The learning rate scheduler ReduceLROnPlateau is applied with a patience of 5 epochs and a decay factor of 0.5. Early stopping is based on EMD loss for \textit{NIMA} and on CCC for \textit{MIR} and \textit{ICI}. \textit{NIMA} and the \textit{pre.} phase of \textit{MIR} and \textit{ICI} are trained with a batch size of 32, while the \textit{fine.} phase of \textit{MIR} and \textit{ICI} uses a batch size of 16. All MLPs use ReLU activations, Kaiming-uniform weight initialization, and a dropout rate of 0.1. The image backbone is CLIP~\cite{radford2021learningtransferablevisualmodels} (ViT-B/16) with OpenAI pretrained weights. Hyperparameters specific to each UDA method were manually tuned based on source-domain behavior only; target-domain data, which would not be observable in practice, was never used in this process. Experiments were conducted on an NVIDIA GeForce RTX 3090 GPU.

\begin{table*}[t]
\centering
\caption{Experimental Results (\textbf{Within-Domain}).}
\vspace{6pt}
\label{tab:within_domain}
\begin{tabular}{c|cc|cc|cc|cc}
\hline
\textbf{Model}
 & \multicolumn{2}{c|}{\textbf{Art (A)}}
 & \multicolumn{2}{c|}{\textbf{Fashion (F)}}
 & \multicolumn{2}{c|}{\textbf{Landscape (L)}}
 & \multicolumn{2}{c}{\textbf{Avg.}} \\
\hline
 & SCC & CCC & SCC & CCC & SCC & CCC & SCC & CCC \\
\hline
\centering GPT 5.4
& .321{\tiny$\pm$.200} & .145{\tiny$\pm$.124} & .211{\tiny$\pm$.154} & .070{\tiny$\pm$.074} & .293{\tiny$\pm$.152} & .154{\tiny$\pm$.097}
& .275 & .123 \\
\centering Claude Opus 4.6
& .327{\tiny$\pm$.180} & .114{\tiny$\pm$.092} & .142{\tiny$\pm$.159} & .039{\tiny$\pm$.058} & .275{\tiny$\pm$.156} & .114{\tiny$\pm$.086}
& .248 & .089 \\
\centering Gemini 3.0 Flash
& .289{\tiny$\pm$.186} & .128{\tiny$\pm$.147} & .156{\tiny$\pm$.159} & .060{\tiny$\pm$.078} & .270{\tiny$\pm$.161} & .183{\tiny$\pm$.120}
& .238 & .124 \\
\hline
\multirow{1}{*}{\centering NIMA \cite{Talebi2018-tj}}
 & .468{\tiny$\pm$.201} & .328{\tiny$\pm$.146} & .337{\tiny$\pm$.164} & .191{\tiny$\pm$.100} & .467{\tiny$\pm$.180} & .324{\tiny$\pm$.142}
 & .424 & .281 \\
\multirow{1}{*}{\centering MIR \cite{zhu2023personalized}}
 & .527{\tiny$\pm$.155} & .486{\tiny$\pm$.152} & .361{\tiny$\pm$.174} & .316{\tiny$\pm$.152} & .483{\tiny$\pm$.176} & .437{\tiny$\pm$.175}
 & .457 & .413 \\
\multirow{1}{*}{\centering ICI \cite{shi2023personalized}}
 & \textbf{\underline{.539}}{\tiny$\pm$.152} & \textbf{\underline{.493}}{\tiny$\pm$.152} & \textbf{\underline{.366}}{\tiny$\pm$.173} & \textbf{\underline{.319}}{\tiny$\pm$.153} & \textbf{\underline{.504}}{\tiny$\pm$.175} & \textbf{\underline{.458}}{\tiny$\pm$.171}
 & \textbf{\underline{.470}} & \textbf{\underline{.423}} \\
 \hline
\end{tabular}
\vspace{-0.3em}
\begin{center}
\small
Values are mean\,$\pm$\,std over 129 users.
SCC: Spearman's rank Correlation Coefficient; CCC: Concordance Correlation Coefficient.
\textbf{Avg.}: mean across the three domains (Art, Fashion, Landscape).
\textbf{\underline{Bold}}: best overall.
\end{center}
\end{table*}

\section{Results and Discussion}
We organize our findings around the central question of whether personalized aesthetic preferences transfer across visual domains. We first establish within-domain PIAA performance as a reference point, comparing the personalized models against the GIAA and proprietary LLM baselines (Sec.~\ref{sec:res-within}). We then evaluate cross-domain transfer under unsupervised domain adaptation (Sec.~\ref{sec:cross_domain}).
\subsection{Within-Domain PIAA}
\label{sec:res-within}
\subsubsection{Results}
Table~\ref{tab:within_domain} reports within-domain performance across the three visual domains. The PIAA models (\textit{ICI}, \textit{MIR}) substantially outperform both the GIAA model (\textit{NIMA}) and the proprietary LLM baselines, with \textit{ICI} achieving the best overall performance (avg.\ SCC=.470, CCC=.423). The three proprietary LLMs---\textit{GPT}, \textit{Claude}, and \textit{Gemini}---perform comparably to one another and, under zero-shot inference, do not reach the performance of \textit{NIMA} in any domain. Across domains, all trained models consistently achieve the highest scores on the art domain and the lowest on fashion. Inter-user variability, as indicated by the standard deviations, is non-negligible across all models, reflecting the substantial heterogeneity of individual aesthetic preferences.
\subsubsection{Discussion}
Three observations are worth highlighting. \textbf{First}, the consistent gap between PIAA models and \textit{NIMA} indicates that incorporating user attributes provides meaningful personalization benefits within each domain, in line with prior findings~\cite{shi2023personalized, zhu2023personalized}. The marginal advantage of \textit{ICI} over \textit{MIR} suggests that graph-based modeling of user--image interactions captures personalized preferences slightly more effectively than the multi-attribute interactive reasoning of \textit{MIR}. 

\textbf{Second}, the limited performance of the proprietary LLMs under zero-shot inference suggests that individual-level aesthetic judgment is difficult to infer from textual user attributes alone in this setting. We note, however, that these models were evaluated with zero-shot inference only, without prompt optimization or in-context (few-shot) examples; their performance may improve with task-specific prompting. These results nonetheless indicate that explicit modeling of user--image interactions from labeled data remains beneficial. 

\textbf{Third}, the cross-domain disparity (art $>$ landscape $>$ fashion) suggests that aesthetic preferences in the fashion domain are intrinsically harder to model. Fashion shows the largest disagreement across annotators, with the lowest ICC(1,1) (=.19; Sec.~\ref{sec:analysis}). Because the GIAA model (\textit{NIMA}) is trained to predict a population-aggregated signal, this disagreement leaves little consistent structure to learn, yielding the weakest GIAA performance on fashion (Table~\ref{tab:within_domain}); since the PIAA models build on this GIAA backbone, the limitation propagates downstream and constrains personalized prediction on fashion as well. This low consensus likely reflects the stronger influence of personal style and cultural context on fashion judgments, which the collected user attributes do not fully capture.

\begin{table*}[t]
\centering
\caption{Experimental Results using ICI (\textbf{Cross-Domain}).}
\vspace{6pt}
\label{tab:cross_domain}
\small
\begin{tabular}{c|cccccc|c}
\hline
\textbf{Method}
 & \textbf{A→F} & \textbf{A→L}
 & \textbf{F→A} & \textbf{F→L}
 & \textbf{L→A} & \textbf{L→F} & \textbf{Avg.} \\
\hline
 & SCC / CCC & SCC / CCC & SCC / CCC & SCC / CCC & SCC / CCC & SCC / CCC & SCC / CCC \\
\hline
 \textcolor{gray}{S. Only}   & \textcolor{gray}{.081 / .047} & \textcolor{gray}{.162 / .087} & \textcolor{gray}{.154 / .082} & \textcolor{gray}{.086 / .044} & \textcolor{gray}{.254 / .127} & \textcolor{gray}{.131 / .083} & \textcolor{gray}{.145 / .078} \\
 \noalign{\vskip 1pt}
\cdashline{1-8}
\noalign{\vskip 1pt}
DANN~\cite{Ganin2016} & .116 / .079 & .164 / .120 & .228 / .143 & .164 / .112 & \underline{.307} / \textbf{\underline{.244}} & .184 / \textbf{\underline{.147}} & .194 / .141 \\
CDAN~\cite{Long2017-tu} & .154 / .103 & .226 / .152 & .083 / .054 & .135 / .085 & .247 / \underline{.198} & .139 / .121 & .164 / .119 \\
ALDA~\cite{chen2020adversariallearnedlossdomainadaptation}    & .087 / .013 & -.026 / -.003 & .103 / .027 & -.010 / .003 & .183 / .060 & .126 / .047 & .077 / .025 \\
DeepJDOT~\cite{damodaran2018deepjdotdeepjointdistribution}    & .153 / \textbf{\underline{.141}} & .284 / \textbf{\underline{.237}} & .101 / .086 & \underline{.202} / \underline{.169} & .038 / .034 & .149 / \underline{.130} & .154 / .132 \\
JUMBOT~\cite{Fatras2021-qw}    & .145 / .120 & .220 / .163 & .104 / .074 & -.024 / -.021 & .236 / .184 & .116 / .084 & .133 / .101 \\
DeepCORAL~\cite{Sun8_35}    & {.209} / \underline{.140} & {.357} / \underline{.217} & {.327} / \textbf{\underline{.257}} & .184 / {.137} & .238 / .110 & .210 / .110 & .254 / \textbf{\underline{.162}} \\
RSD~\cite{Chen2021-ua}    & \underline{.215} / .116 & \textbf{\underline{.362}} / {.163} & \underline{.331} / .228 & \textbf{\underline{.242}} / \textbf{\underline{.174}} & .285 / .155 & \textbf{\underline{.226}} / .121 & \underline{.277} / \underline{.160} \\
DARE-GRAM~\cite{nejjar2023domain} & \textbf{\underline{.219}} / .115 & {\underline{.361}} / .184 & \textbf{\underline{.332}} / \underline{.244} & {\underline{.225}} / .156 & \textbf{\underline{.318}} / .112 & \underline{.225} / \underline{.103} & \textbf{\underline{.280}} / .152 \\
\noalign{\vskip 1pt}
\cdashline{1-8}
\noalign{\vskip 1pt}
 \textcolor{gray}{T. Only} & \textcolor{gray}{.366 / .319} & \textcolor{gray}{.504 / .458} & \textcolor{gray}{.539 / .493} & \textcolor{gray}{.504 / .458} & \textcolor{gray}{.539 / .493} & \textcolor{gray}{.366 / .319} & \textcolor{gray}{.470 / .423} \\
 \hline
\end{tabular}
\vspace{-0.3em}
\begin{center}
Each cell reports SCC / CCC, averaged over 129 users.
\textbf{Avg.}: mean across the six transfer directions.
\textbf{S. Only} (Source-Only): model trained on the source domain and applied to the target without adaptation (lower bound).
\textbf{T. Only} (Target-Only): model trained directly on the target domain (upper bound).
\underline{\textbf{Underlined bold}}: best overall.
\textbf{Bold}: second best overall.
Since DARE-GRAM and RSD are designed specifically for regression tasks, their underlying GIAA model is built upon DeepCORAL, the best-performing UDA method for GIAA.
\end{center}
\end{table*}

\subsection{Cross-Domain PIAA}
\label{sec:cross_domain}
\subsubsection{Results.}
Table~\ref{tab:cross_domain} reports cross-domain PIAA results using \textit{ICI} as the base model under six source--target transfer settings. The Source-Only baseline directly applies a model trained on the source domain to the target domain without any adaptation. It achieves only avg.\ SCC=.145 and CCC=.078, far below the Target-Only upper bound (avg.\ SCC=.470, CCC=.423). This large gap indicates a substantial domain shift in personalized aesthetic preferences and motivates the use of domain adaptation.

Most UDA methods improve upon Source-Only on average, although some (e.g., \textit{ALDA}) fall below it, and the magnitude of improvement varies considerably across method categories. Among adversarial methods, \textit{DANN} and \textit{CDAN} yield moderate gains. Among discrepancy-based methods, optimal-transport-based approaches reach similar levels of performance, while feature-alignment methods perform notably better: \textit{DeepCORAL}, \textit{RSD}, and \textit{DARE-GRAM} form the top tier overall, with the latter two---both explicitly designed for regression tasks---achieving the highest SCC scores among all UDA methods. Performance also varies across transfer directions, but the picture changes depending on whether absolute or relative performance is considered.

In terms of absolute scores, transfers targeting the art domain are consistently among the highest, while those targeting fashion are the lowest. However, when assessed relative to each direction's Target-Only upper bound, the recovery rates are broadly comparable across transfer directions, averaging roughly 60\% of the Target-Only SCC. The main exception is fashion$\rightarrow$landscape, where recovery drops to about 45\%. Overall, the absolute performance gap is largely attributable to the differing difficulty of the target domains themselves, rather than to differences in the adaptability of personalized aesthetic preferences across directions. For CCC, the best methods recover only about 38\% of the Target-Only, compared with roughly 60\% for SCC. This suggests that, in the unsupervised setting, capturing annotator-specific rating tendencies---which CCC reflects beyond rank order---remains difficult.

\subsubsection{Discussion.}
The results above yield two notable insights into the nature of cross-domain personalized aesthetic transfer, as well as a practical outlook. 

\textbf{First}, the clear performance hierarchy across method categories (feature-alignment $>$ adversarial $\approx$ optimal-transport) suggests that the cross-domain PIAA task benefits more from explicit alignment of feature statistics than from adversarial or optimal-transport objectives. The particular effectiveness of \textit{RSD}~\cite{Chen2021-ua} and \textit{DARE-GRAM}~\cite{nejjar2023domain}, both explicitly designed for regression, further indicates that regression-aware UDA strategies are better suited to PIAA than classification-oriented methods.

\textbf{Second}, and most importantly, the fact that the best adaptation methods recover approximately 60\% of the Target-Only SCC upper bound on average---roughly twice the Source-Only baseline---under a fully unsupervised setting constitutes encouraging evidence that personalized aesthetic preferences are, to a meaningful extent, transferable across visual domains. This supports the hypothesis that individuals carry domain-invariant aesthetic dispositions alongside domain-specific ones. At the same time, the remaining gap to the Target-Only upper bound indicates that current adaptation techniques do not fully exploit the structure of personalized aesthetic preferences. We view this as an opportunity rather than a limitation: it motivates the development of PIAA-specific adaptation methods that explicitly model the interplay between user attributes and domain shift, which we leave for future work.

\textbf{Third}, while the top-tier UDA methods currently perform only on par with the proprietary LLMs, we believe this margin can be widened in their favor---both by scaling up the training data (e.g., with publicly available PIAA datasets) and by developing task-specific method improvements.

\begin{figure*}[t]
    \centering
    \includegraphics[width=1\linewidth]{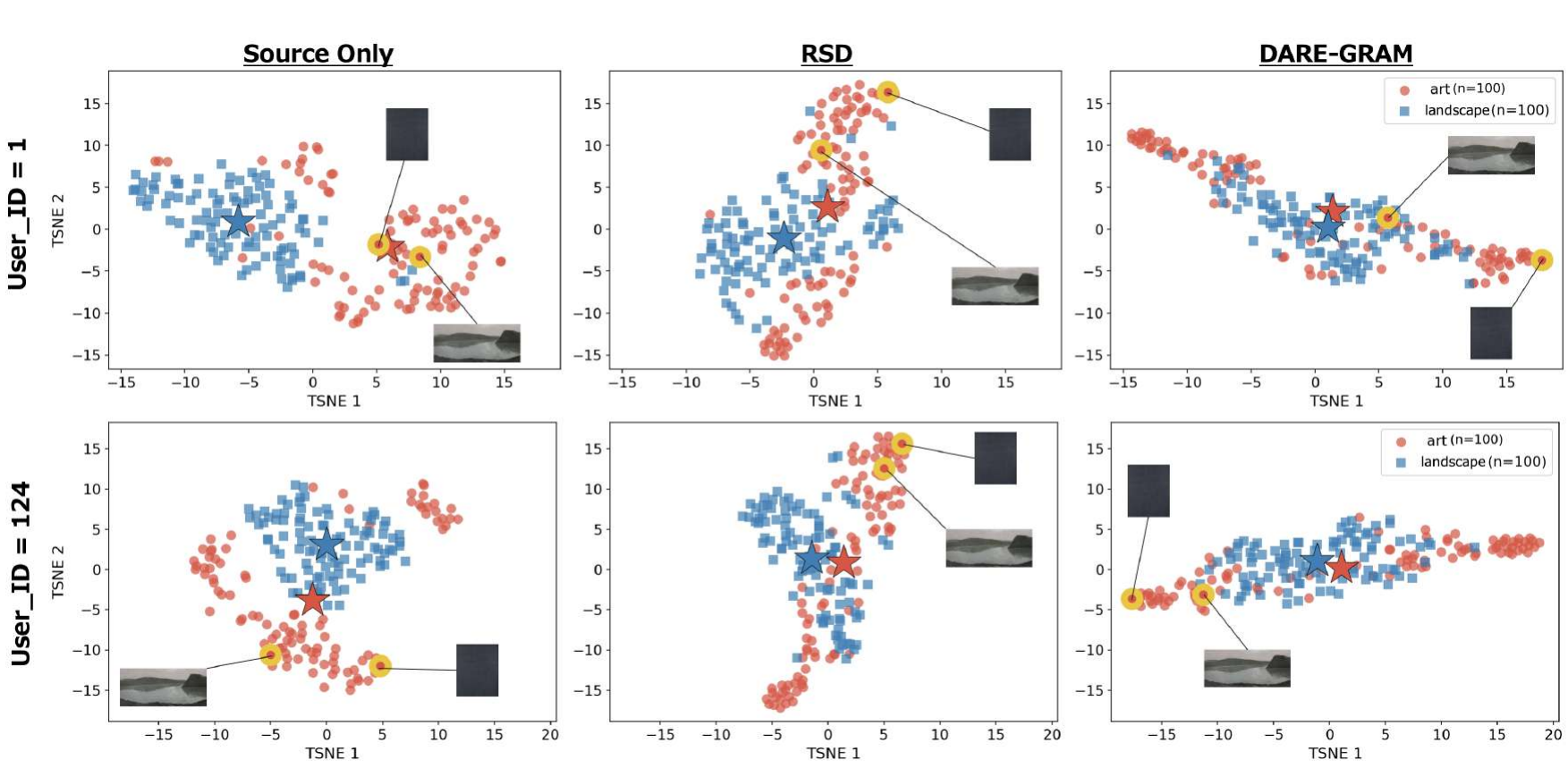}
    \vspace{0.2em}
\caption{t-SNE visualization of intermediate representations learned by ICI (\textit{fine.} phase) for target users (User IDs 1 and 124) under the A$\rightarrow$L setting. Red circles and blue squares denote art and landscape stimuli; stars mark per-domain centroids. Highlighted points indicate a near-black abstract artwork and a dark landscape painting.}
    \label{fig:tsne}
\end{figure*}

\subsection{How Domain Adaptation Reshapes the Feature Space}
\label{sec:res-feature}
To complement the quantitative results in Sec.~\ref{sec:cross_domain}, we qualitatively inspect the intermediate representations learned by \textit{ICI} under the A$\rightarrow$L transfer setting. This direction was chosen because it achieves the highest recovery rate (SCC=71.6\%) among the six transfer settings. After \textit{fine.}-phase training, we extract features from the last hidden layer of the user--stimulus interaction module for target users (User IDs 1 and 124) and project them into two dimensions using t-SNE (Fig.~\ref{fig:tsne}). We selected these two users because they form one of the very few pairs in XPASS-Vis who rated the same two stimuli (shown in Fig.~\ref{fig:tsne}). We compare three conditions: the Source-Only baseline and two top-tier discrepancy-based methods, \textit{RSD} and \textit{DARE-GRAM}.

Three patterns are visible in Fig.~\ref{fig:tsne}. \textbf{First}, the distance between the per-domain centroids (stars) shrinks markedly after applying UDA, which is consistent with the marginal-alignment objective shared by both methods. \textbf{Second}, focusing on a visually similar pair of \emph{dark} stimuli---a near-black abstract artwork and a dark landscape painting---that happen to be embedded near each other under Source-Only, we observe that after adaptation the dark abstract artwork moves further from the landscape cluster, while the dark landscape painting moves closer to it. \textbf{Third}, despite the centroid shift, the two clusters remain only partially overlapping under both \textit{RSD} and \textit{DARE-GRAM}, with a substantial fraction of art and landscape points still occupying distinct regions of the embedding space.

These observations suggest that the UDA objectives do not merely shrink the global feature distance in a content-agnostic way, but rather reorganize the representation according to semantic content: stimuli are repositioned by their semantic category (e.g., the dark landscape painting moving toward the landscape cluster) rather than by low-level visual appearance. We note, however, that this pair was selected for illustrative purposes and should be read as anecdotal rather than systematic evidence. Furthermore, whereas successful adaptation on standard benchmarks such as digit recognition typically yields near-complete mixing of source and target features, the persistent separation observed here likely reflects the inherent difficulty of the cross-domain PIAA task: aesthetic preferences are tied to high-level semantic and stylistic attributes that differ substantially across domains, making alignment fundamentally harder than aligning low-level visual statistics. This interpretation is consistent with the quantitative finding in Sec.~\ref{sec:cross_domain} that the best UDA methods recover only $\sim$60\% of the Target-Only SCC, motivating PIAA-specific adaptation strategies as future work.

\begin{table*}[t]
\centering
\caption{Top-5 features per domain$\to$domain pair, ranked by $|\beta_{\mathrm{std}}|$ from OLS on ICI (DARE-GRAM, SCC). }
\label{tab:da_factors_top5}
\setlength{\tabcolsep}{4pt}
\renewcommand{\arraystretch}{1.05}
\footnotesize
\begin{tabular}{rlrr@{\hspace{8pt}}rlrr}
\toprule
\multicolumn{4}{c}{\textbf{art$\to$fashion}} & \multicolumn{4}{c}{\textbf{art$\to$landscape}} \\
\midrule
\# & feature & $\beta_{\mathrm{std}}$ & $p_{\mathrm{FDR}}$ & \# & feature & $\beta_{\mathrm{std}}$ & $p_{\mathrm{FDR}}$ \\
\midrule
1 & \texttt{baseline\_scc\_tgt} & $-.073$$^{\dagger}$ & .000 & 1 & \texttt{baseline\_scc\_tgt} & $-.115$$^{\dagger}$ & .000 \\
2 & \texttt{shift\_std} & $+.028$ & .934 & 2 & \texttt{gender\_male} & $-.042$ & .153 \\
3 & \texttt{big5\_C} & $+.022$ & .934 & 3 & \texttt{big5\_C} & $-.038$ & .153 \\
4 & \texttt{shift\_interest} & $-.018$ & .934 & 4 & \texttt{nationality\_JPN} & $+.036$ & .153 \\
5 & \texttt{big5\_A} & $-.017$ & .934 & 5 & \texttt{baseline\_scc\_src} & $+.033$ & .153 \\
\midrule  \midrule
\multicolumn{4}{c}{\textbf{fashion$\to$art}} & \multicolumn{4}{c}{\textbf{fashion$\to$landscape}} \\
 \midrule
\# & feature & $\beta_{\mathrm{std}}$ & $p_{\mathrm{FDR}}$ & \# & feature & $\beta_{\mathrm{std}}$ & $p_{\mathrm{FDR}}$ \\
\midrule
1 & \texttt{baseline\_scc\_tgt} & $-.113$$^{\dagger}$ & .000 & 1 & \texttt{baseline\_scc\_tgt} & $-.135$$^{\dagger}$ & .000 \\
2 & \texttt{baseline\_scc\_src} & $+.040$ & .455 & 2 & \texttt{nationality\_JPN} & $+.071$$^{\dagger}$ & .007 \\
3 & \texttt{gender\_other} & $-.035$ & .455 & 3 & \texttt{gender\_male} & $-.063$ & .035 \\
4 & \texttt{shift\_retest\_mae} & $-.031$ & .457 & 4 & \texttt{big5\_ES} & $+.050$ & .079 \\
5 & \texttt{big5\_E} & $+.029$ & .499 & 5 & \texttt{shift\_skew} & $-.047$ & .102 \\
\midrule
 \midrule
\multicolumn{4}{c}{\textbf{landscape$\to$art}} & \multicolumn{4}{c}{\textbf{landscape$\to$fashion}} \\
 \midrule
\# & feature & $\beta_{\mathrm{std}}$ & $p_{\mathrm{FDR}}$ & \# & feature & $\beta_{\mathrm{std}}$ & $p_{\mathrm{FDR}}$ \\
\midrule
1 & \texttt{baseline\_scc\_tgt} & $-.151$$^{\dagger}$ & .000 & 1 & \texttt{baseline\_scc\_tgt} & $-.046$$^{\dagger}$ & .002 \\
2 & \texttt{shift\_generality} & $-.046$ & .167 & 2 & \texttt{shift\_kurt} & $+.026$ & .506 \\
3 & \texttt{big5\_ES} & $+.033$ & .598 & 3 & \texttt{nationality\_JPN} & $+.020$ & .541 \\
4 & \texttt{shift\_mean} & $+.029$ & .623 & 4 & \texttt{baseline\_scc\_src} & $+.018$ & .541 \\
5 & \texttt{gender\_male} & $+.022$ & .729 & 5 & \texttt{big5\_E} & $+.018$ & .541 \\
\midrule
 \midrule
\multicolumn{8}{c}{\textbf{Aggregated mean across 6 pairs}} \\
\midrule
\# & feature & $\overline{\beta}\,\pm\,\mathrm{SD}$ & $n_{\mathrm{sig}}/n_{\mathrm{pairs}}$ & \# & feature & $\overline{\beta}\,\pm\,\mathrm{SD}$ & $n_{\mathrm{sig}}/n_{\mathrm{pairs}}$ \\
\midrule
1 & \texttt{baseline\_scc\_tgt} & $-.105${\scriptsize$\,\pm\,.039$} & 6/6 & 6 & \texttt{shift\_generality} & $-.019${\scriptsize$\,\pm\,.018$} & 0/6 \\
2 & \texttt{nationality\_JPN} & $+.017${\scriptsize$\,\pm\,.034$} & 1/6 & 7 & \texttt{shift\_mean} & $+.010${\scriptsize$\,\pm\,.019$} & 0/6 \\
3 & \texttt{gender\_male} & $-.017${\scriptsize$\,\pm\,.031$} & 1/6 & 8 & \texttt{shift\_skew} & $-.016${\scriptsize$\,\pm\,.018$} & 0/6 \\
4 & \texttt{big5\_ES} & $+.024${\scriptsize$\,\pm\,.020$} & 0/6 & 9 & \texttt{edu\_level} & $+.018${\scriptsize$\,\pm\,.009$} & 0/6 \\
5 & \texttt{baseline\_scc\_src} & $+.018${\scriptsize$\,\pm\,.018$} & 0/6 & 10 & \texttt{big5\_E} & $+.005${\scriptsize$\,\pm\,.020$} & 0/6 \\
\bottomrule
\end{tabular}
\vspace{-0.6em}
\begin{center}
$^{\dagger}p_{\mathrm{FDR}}<.01$.
The bottom block reports the mean\,$\pm$\,SD of $\beta_{\mathrm{std}}$ across all six pairs and the number of pairs ($n_{\mathrm{sig}}/n_{\mathrm{pairs}}$) in which each feature is FDR-significant ($p_{\mathrm{FDR}}<.01$).
All variance inflation factors (VIFs) were below 2.2, indicating that multicollinearity was not a concern.
\end{center}
\end{table*}

\subsection{User-Level Analysis of Domain Adaptation}
\label{sec:who_gain}
The substantial inter-user variability observed in Table~\ref{tab:within_domain} raises a complementary question: \emph{which users benefit from domain adaptation, and which do not?}

\subsubsection{Analysis Setup.}
We define the per-user UDA improvement on a source$\to$target pair $(s,t)$ as
\begin{equation}
\Delta_u^{s\to t} \;=\; \mathrm{SCC}_u^{\mathrm{DA}}(t) - \mathrm{SCC}_u^{\mathrm{noDA}}(t),
\end{equation}
where $\mathrm{SCC}_u^{\mathrm{noDA}}(t)$ denotes the SCC of the Source-Only model and $\mathrm{SCC}_u^{\mathrm{DA}}(t)$ that of the \textit{DARE-GRAM}-adapted model, which is the best-performing UDA method on average in our benchmark.
We regress $\Delta_u$ on a set of standardized user-level features comprising:
\begin{itemize}
\item demographics (age, education, gender, nationality)
\item the BigFive personality factors
\item per-domain formal-learning indicators
\item the absolute target--source difference (\texttt{shift\_*}) of each per-domain statistic: the per-user mean, standard deviation, skewness, and kurtosis of overall aesthetic ratings; self-reported domain interest; test--retest MAE; and \texttt{generality}, defined as the Pearson correlation between a user's ratings and the leave-one-out mean of other annotators on shared images (capturing how ``typical'' their taste is)
\item the Source-Only model's source- and target-domain SCC (\texttt{baseline\_scc\_src}, \texttt{baseline\_scc\_tgt}), included only as controls for regression-to-the-mean.
\end{itemize}
Because the source- and target-domain levels of each per-domain statistic and their difference were strongly collinear, we retained only the absolute target--source differences (\texttt{shift\_*}); after this reduction all variance inflation factors were below 2.2 (Table~\ref{tab:da_factors_top5}). We further note that \texttt{baseline\_scc\_tgt} is mechanically (negatively) correlated with $\Delta_u$, since $\Delta_u$ subtracts $\mathrm{SCC}_u^{\mathrm{noDA}}(t)$ by construction; we therefore include it purely as a control and do not interpret its coefficient substantively.

With all features z-standardized, each coefficient $\beta_{\mathrm{std}}$ represents the change in $\Delta_u$ associated with a $+1$~SD change in that feature. We fit OLS regressions independently for each of the six $s\to t$ pairs, apply Benjamini--Hochberg FDR correction~\cite{Benjamini1995-jk} to the $p$-values, and aggregate across pairs. For each feature we report the cross-pair mean $\overline{\beta}$, its SD, and the number of pairs $n_{\mathrm{sig}}$ in which $p_{\mathrm{FDR}}<.01$ (Table~\ref{tab:da_factors_top5}).

\subsubsection{Results and Discussion.}
Only one feature is associated with $\Delta_u$ across all six transfer directions: \texttt{baseline\_scc\_tgt} ($\overline{\beta}=-.105 \pm .051$, significant in 6/6 pairs). As noted above, however, this negative association is a mechanical consequence of the difference-based outcome (regression-to-the-mean) rather than a behavioral property of users, and we do not interpret it. Once this control is set aside, no user-level feature is FDR-significant in more than one transfer direction: only \texttt{nationality\_JPN} and \texttt{gender\_male} reach significance, each in a single pair. All remaining per-domain difference statistics, demographic variables, and Big Five traits yield $|\overline{\beta}|<.025$ in the aggregation, indicating idiosyncratic rather than systematic effects.

This analysis yields a clear conclusion that we read as encouraging rather than negative: apart from the mechanical baseline term, none of the user attributes collected in XPASS-Vis (demographics, personality, rating-style statistics, or generality) systematically predicts who benefits from domain adaptation. The few significant effects are confined to individual transfer directions, and the remaining coefficients are not merely non-significant but small in magnitude ($|\overline{\beta}|<.025$). Rather than a failure to find structure, this indicates that the gains of marginal-alignment UDA are broad-based: they are not concentrated in, or gated by, any attribute-defined subgroup, and the method does not exploit coarse demographic or personality shortcuts but instead leverages finer, individual-level preference signal. Whether a given user benefits therefore appears to be governed by something beyond static traits---plausibly an intra-individual, cross-domain property: how much that user's preference structure is shared between the source and target domains. This does leave one practical limitation: being able to anticipate a priori which users will benefit would be valuable for reducing adaptation cost---applying transfer only to users for whom it is likely to help---but the attribute-level analysis used here does not reach the information required for such prospective triage. Because the relevant cross-domain preference overlap is not captured by the attributes examined here, recovering it will require more refined analyses---for example, directly quantifying the per-user consistency of preference structure across domains---than the attribute-level regression used in this study. We view this as the central methodological challenge for PIAA-specific UDA and leave it to future work.

\section{LIMITATION}
\label{sec:limitation}
While XPASS-Vis and our benchmark provide a foundation for cross-domain PIAA research, several limitations remain. 
\textbf{First}, like most existing PIAA datasets, the annotator pool is geographically and culturally skewed: almost all participants are from East Asia, and very few reported formal education in the evaluated domains. Aesthetic preferences are known to be influenced by cultural background, and the extent to which the patterns observed in XPASS-Vis generalize to other populations remains an open question.

\textbf{Second}, the scope of domains covered is limited in two respects. Within the visual modality, the three domains---art, fashion, and landscape---represent only a subset of everyday aesthetic experience, and other visual domains such as architecture or product design were not included. More broadly, XPASS-Vis is restricted to the visual modality, whereas aesthetic experience naturally spans multiple modalities. In addition, the landscape domain originally consists of video clips, but we used single sampled frames for modality consistency and computational efficiency, discarding temporal information that may contribute to aesthetic judgment. Extending cross-domain PIAA to cross-modal settings (e.g., fashion$\rightarrow$music) is an important future direction. 

\textbf{Third}, our benchmark focuses on unsupervised domain adaptation as a first step, but other transfer settings---such as few-shot adaptation and source-free adaptation---are equally relevant to cross-domain PIAA and remain unexplored. Moreover, extending these methods to address PIAA-specific challenges is an important direction that we leave for future research.

\section{Conclusion}
This work asked whether personalized aesthetic preferences can be transferred across visual domains, and what limits such transfer. Our answer is a qualified yes. On XPASS-Vis, the first dataset in which the same annotators rate multiple domains with sufficient per-domain samples, the best unsupervised adaptation methods recover roughly 60\% of the Target-Only upper bound \emph{without any target-domain labels}---evidence that individuals carry domain-invariant aesthetic dispositions alongside domain-specific ones.

This transfer is not supported equally by all strategies: regression-aware feature-alignment methods (\textit{RSD}, \textit{DARE-GRAM}) consistently outperform adversarial and optimal-transport approaches, as they better suit the continuous nature of aesthetic scores. Crucially, our per-user analysis asked \emph{who} benefits, and the answer is encouraging: apart from a mechanical regression-to-the-mean effect, no user attribute in XPASS-Vis---demographics, personality, rating-style statistics, or how typical a user's taste is---predicts the gain, and all effects are small. The benefit of marginal-alignment UDA is thus broad-based rather than confined to any attribute-defined subgroup, suggesting it is driven by a finer, intra-individual property---how much a user's preference structure is shared across domains---than static traits capture. The central challenge for PIAA-specific adaptation is therefore to characterize and exploit this cross-domain preference overlap---also a prerequisite for anticipating benefit a priori for cost-saving user triage---which we leave to future work alongside extensions to broader transfer settings, visual domains, and cross-modal scenarios.


\bibliographystyle{unsrt}  
\bibliography{references}  

@String(CVPR= {IEEE Conf. Comput. Vis. Pattern Recog.})

@String(ICCV= {Int. Conf. Comput. Vis.})

@String(ECCV= {Eur. Conf. Comput. Vis.})

@String(AAAI = {AAAI})

@String(CVPR  = {CVPR})

@String(ICCV  = {ICCV})

@String(ECCV  = {ECCV})

@ARTICLE{Talebi2018-tj,
  title     = "{NIMA}: Neural Image Assessment",
  author    = "Talebi, Hossein and Milanfar, Peyman",
  journal   = "IEEE Trans. Image Process.",
  publisher = "Institute of Electrical and Electronics Engineers (IEEE)",
  volume    =  27,
  number    =  8,
  pages     = "3998--4011",
  month     =  aug,
  year      =  2018
}

@misc{chen2025role,
      title={On the Role of Individual Differences in Current Approaches to Computational Image Aesthetics}, 
      author={Li-Wei Chen and Ombretta Strafforello and Anne-Sofie Maerten and Tinne Tuytelaars and Johan Wagemans},
      year={2025},
      eprint={2502.20518},
      archivePrefix={arXiv},
      primaryClass={cs.CV},
      url={https://arxiv.org/abs/2502.20518}, 
}

@inproceedings{nejjar2023domain,
  title={DARE-GRAM : Unsupervised Domain Adaptation Regression by Aligning Inversed Gram Matrices},
  author={Nejjar, Ismail and Wang, Qin and Fink, Olga},
  booktitle={Proceedings of the IEEE/CVF Conference on Computer Vision and Pattern Recognition.},
  year={2023}
}

@inproceedings{damodaran2018deepjdotdeepjointdistribution,
author = {Damodaran, Bharath Bhushan and Kellenberger, Benjamin and Flamary, R\'{e}mi and Tuia, Devis and Courty, Nicolas},
title = {DeepJDOT: Deep Joint Distribution Optimal Transport for Unsupervised Domain Adaptation},
year = {2018},
booktitle = {Computer Vision – ECCV 2018: 15th European Conference, Munich, Germany, September 8-14, 2018, Proceedings, Part IV},
pages = {467–483},
numpages = {17},
}

@ARTICLE{Benjamini1995-jk,
  title     = "Controlling the false discovery rate: A practical and powerful
               approach to multiple testing",
  author    = "Benjamini, Yoav and Hochberg, Yosef",
  journal   = "J. R. Stat. Soc. Series B Stat. Methodol.",
  publisher = "Oxford University Press (OUP)",
  volume    =  57,
  number    =  1,
  pages     = "289--300",
  month     =  jan,
  year      =  1995,
  language  = "en"
}

@misc{radford2021learningtransferablevisualmodels,
      title={Learning Transferable Visual Models From Natural Language Supervision}, 
      author={Alec Radford and Jong Wook Kim and Chris Hallacy and Aditya Ramesh and Gabriel Goh and Sandhini Agarwal and Girish Sastry and Amanda Askell and Pamela Mishkin and Jack Clark and Gretchen Krueger and Ilya Sutskever},
      year={2021},
      eprint={2103.00020},
      archivePrefix={arXiv},
      primaryClass={cs.CV},
      url={https://arxiv.org/abs/2103.00020}, 
}

@INPROCEEDINGS{Yang2019-ji,
  title     = "Outlier detection: how to threshold outlier scores?",
  author    = "Yang, Jiawei and Rahardja, Susanto and Fränti, Pasi",
  booktitle = "Proceedings of the International Conference on Artificial
               Intelligence, Information Processing and Cloud Computing",
  publisher = "ACM",
  address   = "New York, NY, USA",
  month     =  dec,
  year      =  2019
}

@inproceedings{kong2016photoaestheticsrankingnetwork,
  title={Photo Aesthetics Ranking Network with Attributes and Content Adaptation},
  author={Kong, Shu and Shen, Xiaohui and Lin, Zhe and Mech, Radomir and Fowlkes, Charless},
  booktitle={ECCV},
  year={2016}
}

@misc{behrad2025charmmissingpiecevit,
      title={Charm: The Missing Piece in ViT fine-tuning for Image Aesthetic Assessment}, 
      author={Fatemeh Behrad and Tinne Tuytelaars and Johan Wagemans},
      year={2025},
      eprint={2504.02522},
      archivePrefix={arXiv},
      primaryClass={cs.CV},
      url={https://arxiv.org/abs/2504.02522}, 
}

@misc{wang2026enhancingzeroshotpersonalizedimage,
      title={Enhancing Zero-shot Personalized Image Aesthetics Assessment with Profile-aware Multimodal LLM}, 
      author={Chun Wang and Chenfeng Wei and Chenyang Liu and Weihong Deng},
      year={2026},
      eprint={2604.17233},
      archivePrefix={arXiv},
      primaryClass={cs.CV},
      howpublished  = {arXiv preprint arXiv:2604.17233 [cs.CV]},
      url={https://arxiv.org/abs/2604.17233}, 
}

@misc{ryu2026vl,
      title={What Do Vision-Language Models Encode for Personalized Image Aesthetics Assessment?}, 
      author={Koki Ryu and Hitomi Yanaka},
      year={2026},
      eprint={2604.11374},
      archivePrefix={arXiv},
      primaryClass={cs.CV},
      howpublished={arXiv preprint arXiv:2604.11374},
      url={https://arxiv.org/abs/2604.11374}, 
}

@inproceedings{He_ijcai2022p132,
  title     = {Rethinking Image Aesthetics Assessment: Models, Datasets and Benchmarks},
  author    = {He, Shuai and Zhang, Yongchang and Xie, Rui and Jiang, Dongxiang and Ming, Anlong},
  booktitle = {Proceedings of the Thirty-First International Joint Conference on
               Artificial Intelligence, {IJCAI-22}},
  publisher = {International Joint Conferences on Artificial Intelligence Organization},
  editor    = {Lud De Raedt},
  pages     = {942--948},
  year      = {2022},
  month     = {7},
  note      = {Main Track},
  doi       = {10.24963/ijcai.2022/132},
  url       = {https://doi.org/10.24963/ijcai.2022/132},
}

@InProceedings{yi2023artisticimageaestheticsassessment,
    author    = {Yi, Ran and Tian, Haoyuan and Gu, Zhihao and Lai, Yu-Kun and Rosin, Paul L.},
    title     = {Towards Artistic Image Aesthetics Assessment: A Large-Scale Dataset and a New Method},
    booktitle = {Proceedings of the IEEE/CVF Conference on Computer Vision and Pattern Recognition (CVPR)},
    month     = {June},
    year      = {2023},
    pages     = {22388-22397}
}

@inproceedings{chen2020adversariallearnedlossdomainadaptation,
  title     = {Adversarial-Learned Loss for Domain Adaptation},
  author    = {Chen, Minghao and Zhao, Shuai and Liu, Haifeng and Cai, Deng},
  booktitle = {Proceedings of the AAAI Conference on Artificial Intelligence},
  volume    = {34},
  number    = {04},
  pages     = {3521--3528},
  year      = {2020},
}

@InProceedings{Fatras2021-qw,
  title = 	 {Unbalanced minibatch Optimal Transport; applications to Domain Adaptation},
  author =       {Fatras, Kilian and Sejourne, Thibault and Flamary, R{\'e}mi and Courty, Nicolas},
  booktitle = 	 {Proceedings of the 38th International Conference on Machine Learning},
  pages = 	 {3186--3197},
  year = 	 {2021},
  editor = 	 {Meila, Marina and Zhang, Tong},
  volume = 	 {139},
  series = 	 {Proceedings of Machine Learning Research},
  month = 	 {18--24 Jul},
  publisher =    {PMLR},
  url = 	 {https://proceedings.mlr.press/v139/fatras21a.html}
}

@article {Pham2026_aes,
	author = {Pham, Trung Quang and Chikazoe, Junichi},
	title = {Cross Domain Consistency of Aesthetic Preference-driven Social Behavior},
	elocation-id = {2026.03.21.713367},
	year = {2026},
	doi = {10.64898/2026.03.21.713367},
	publisher = {Cold Spring Harbor Laboratory},
	URL = {https://www.biorxiv.org/content/early/2026/03/25/2026.03.21.713367},
	eprint = {https://www.biorxiv.org/content/early/2026/03/25/2026.03.21.713367.full.pdf},
	journal = {bioRxiv}
}

@inproceedings{Long2017-tu,
author = {Long, Mingsheng and Cao, Zhangjie and Wang, Jianmin and Jordan, Michael I.},
title = {Conditional adversarial domain adaptation},
year = {2018},
booktitle = {Proceedings of the 32nd International Conference on Neural Information Processing Systems},
pages = {1647–1657},
numpages = {11},
}

@ARTICLE{Chen2021-ua,
  title    = "Representation subspace distance for domain adaptation regression",
  author   = "Chen, Xinyang and Wang, Sinan and Wang, Jianmin and Long,
              Mingsheng",
  journal  = "ICML",
  pages    = "1749--1759",
  year     =  2021
}

@InProceedings{Sun8_35,
author="Sun, Baochen
and Saenko, Kate",
editor="Hua, Gang
and J{\'e}gou, Herv{\'e}",
title="Deep CORAL: Correlation Alignment for Deep Domain Adaptation",
booktitle="Computer Vision -- ECCV 2016 Workshops",
year="2016",
publisher="Springer International Publishing",
address="Cham",
pages="443--450",
isbn="978-3-319-49409-8"
}

@article{Ganin2016,
author = {Ganin, Yaroslav and Ustinova, Evgeniya and Ajakan, Hana and Germain, Pascal and Larochelle, Hugo and Laviolette, Fran\c{c}ois and Marchand, Mario and Lempitsky, Victor},
title = {Domain-adversarial training of neural networks},
year = {2016},
issue_date = {January 2016},
publisher = {JMLR.org},
volume = {17},
number = {1},
issn = {1532-4435},
journal = {J. Mach. Learn. Res.},
month = jan,
pages = {2096–2030},
numpages = {35}
}

@ARTICLE{Jacobsen2017-lf,
  title     = "Domain generality and domain specificity in aesthetic
               appreciation",
  author    = "Jacobsen, Thomas and Beudt, Susan",
  journal   = "New Ideas Psychol.",
  publisher = "Elsevier BV",
  volume    =  47,
  pages     = "97--102",
  month     =  dec,
  year      =  2017,
  language  = "en"
}

@ARTICLE{Gosling2003-ct,
  title     = "A very brief measure of the Big-Five personality domains",
  author    = "Gosling, Samuel D and Rentfrow, Peter J and Swann, Jr, William B",
  journal   = "J. Res. Pers.",
  publisher = "Elsevier BV",
  volume    =  37,
  number    =  6,
  pages     = "504--528",
  month     =  dec,
  year      =  2003,
  language  = "en"
}

@ARTICLE{Oshio2012-bf,
  title     = "Development, reliability, and validity of the Japanese version of
               ten item personality inventory ({TIPI}-{J})",
  author    = "Oshio, Atsushi and Abe, Shingo and Cutrone, Pino",
  journal   = "Pasonariti Kenkyu",
  publisher = "Japan Society of Personality Psychology",
  volume    =  21,
  number    =  1,
  pages     = "40--52",
  year      =  2012,
  language  = "en"
}

@inproceedings{ren2017personalized,
  title={Personalized image aesthetics},
  author={Ren, Jie and Lee, Yilin and Chen, Michael and Wang, Xinlei and Fang, Chen},
  booktitle={ICCV},
  pages={638--647},
  year={2017}
}

@InProceedings{Zhong_2025_CVPR,
    author    = {Zhong, Haobin and He, Shuai and Ming, Anlong and Ma, Huadong},
    title     = {Rethinking Personalized Aesthetics Assessment: Employing Physique Aesthetics Assessment as An Exemplification},
    booktitle = {Proceedings of the IEEE/CVF Conference on Computer Vision and Pattern Recognition (CVPR)},
    year      = {2025},
    pages     = {2935-2944}
}

@ARTICLE{Redies2025-wq,
  title     = "A toolbox for calculating quantitative image properties in
               aesthetics research",
  author    = "Redies, Christoph and Bartho, Ralf and Koßmann, Lisa and Spehar,
               Branka and Hübner, Ronald and Wagemans, Johan and
               Hayn-Leichsenring, Gregor U",
  journal   = "Behav. Res. Methods",
  publisher = "Springer Science and Business Media LLC",
  volume    =  57,
  number    =  4,
  pages     =  117,
  month     =  mar,
  year      =  2025,
  language  = "en"
}

@inproceedings{murray2012ava,
  title={{AVA: A large-scale database for aesthetic visual analysis}},
  author={Murray, Naila and Marchesotti, Luca and Perronnin, Florent},
  booktitle={Proceedings of the IEEE Conference on Computer Vision and Pattern Recognition (CVPR)},
  pages={2408--2415},
  year={2012}
}

@misc{yang2026finegrainedigiaa,
      title={Fine-grained Image Aesthetic Assessment: Learning Discriminative Scores from Relative Ranks}, 
      author={Zhichao Yang and Jianjie Wang and Zhixianhe Zhang and Pangu Xie and Xiangfei Sheng and Pengfei Chen and Leida Li},
      year={2026},
      eprint={2603.03907},
      archivePrefix={arXiv},
      primaryClass={cs.CV},
      url={https://arxiv.org/abs/2603.03907}, 
}

@inproceedings{nieto2022understandingaestheticslanguagephoto,
    title={Understanding Aesthetics with Language: A Photo Critique Dataset for Aesthetic Assessment},
    author={Daniel Vera Nieto and Luigi Celona and Clara Fernandez Labrador},
    booktitle={Thirty-sixth Conference on Neural Information Processing Systems Datasets and Benchmarks Track},
    year={2022},
    url={https://openreview.net/forum?id=-VyJim9UBxQ}
}

@inproceedings{yang2022personalizedimageaestheticsassessment,
    author = {Yang, Yuzhe and Xu, Liwu and Li, Leida and Qie, Nan and Li, Yaqian and Zhang, Peng and Guo, Yandong},
    year = {2022},
    month = {06},
    pages = {19829-19837},
    booktitle={Proceedings of the IEEE/CVF Conference on Computer Vision and Pattern Recognition (CVPR)},
    title = {Personalized Image Aesthetics Assessment with Rich Attributes},
    doi = {10.1109/CVPR52688.2022.01924}
}

@Inproceedings{zhang2021lapis,
  title     = {LAPIS: a novel dataset for personalized image aesthetic assessment},
  author    = {Anne-Sofie Maerten and Li-Wei Chen and Stefanie De Winter and Christophe Bossens and Johan Wagemans},
  booktitle = {Proceedings of the IEEE/CVF Conference on Computer Vision and Pattern Recognition (CVPR) Workshops},
  year      = {2025},
}

@inproceedings{yang2014clothing,
  title={Clothing Co-Parsing by Joint Image Segmentation and Labeling},
  author={Yang, Wei and Luo, Ping and Lin, Liang and Wang, Xiaogang},
  booktitle={Proceedings of the IEEE Conference on Computer Vision and Pattern Recognition (CVPR)},
  pages={3182--3189},
  year={2014}
}

@ARTICLE{Schindler2017-kz,
  title     = "Measuring aesthetic emotions: A review of the literature and a
               new assessment tool",
  author    = "Schindler, Ines and Hosoya, Georg and Menninghaus, Winfried and
               Beermann, Ursula and Wagner, Valentin and Eid, Michael and
               Scherer, Klaus R",
  journal   = "PLoS One",
  publisher = "Public Library of Science (PLoS)",
  volume    =  12,
  number    =  6,
  pages     = "e0178899",
  month     =  jun,
  year      =  2017,
  language  = "en"
}

@article{zhu2023personalized,
    author = {Zhu, Hancheng and Yong, Zhou and Shao, Zhiwen and Du, Wenliang and Wang, Guangcheng and Li, Qiaoyue},
    year = {2022},
    month = {11},
    pages = {4181},
    title = {Personalized Image Aesthetics Assessment via Multi-Attribute Interactive Reasoning},
    volume = {10},
    journal = {Mathematics},
    doi = {10.3390/math10224181}
}

@article{shi2023personalized,
title = {Personalized Image Aesthetics Assessment based on Graph Neural Network and Collaborative Filtering},
journal = {Knowledge-Based Systems},
volume = {294},
pages = {111749},
year = {2024},
issn = {0950-7051},
doi = {https://doi.org/10.1016/j.knosys.2024.111749},
url = {https://www.sciencedirect.com/science/article/pii/S0950705124003848},
author = {Huiying Shi and Jing Guo and Yongzhen Ke and Kai Wang and Shuai Yang and Fan Qin and Liming Chen}
}

@article{li2025sekai,
    title={Sekai: A Video Dataset towards World Exploration}, 
    author={Zhen Li and Chuanhao Li and Xiaofeng Mao and Shaoheng Lin and Ming Li and Shitian Zhao and Zhaopan Xu and Xinyue Li and Yukang Feng and Jianwen Sun and Zizhen Li and Fanrui Zhang and Jiaxin Ai and Zhixiang Wang and Yuwei Wu and Tong He and Jiangmiao Pang and Yu Qiao and Yunde Jia and Kaipeng Zhang},
    journal={arXiv preprint arXiv:2506.15675},
    year={2025}
}

\clearpage
\appendix
\renewcommand{\thefigure}{S\arabic{figure}}
\renewcommand{\thetable}{S\arabic{table}}
\setcounter{figure}{0}
\setcounter{table}{0}

\begin{center}
{\LARGE\bfseries Appendix}
\end{center}

\section{Hyperparameter Selection for Unsupervised Domain Adaptation}
\label{sec:uda_hparam}
Comparing adaptation methods fairly requires a principled and consistent hyperparameter selection protocol, especially since these methods were originally designed for classification. While each method requires its own considerations, we adopt three common principles. \textbf{First}, we require that the adaptation objective not substantially degrade learning on the source domain, since a method whose adaptation loss overwhelms the supervised regression loss cannot produce a predictor that is useful for either domain, regardless of how well the feature distributions are aligned. \textbf{Second}, we follow the values reported in the original papers; however, because those values were tuned for cross-entropy loss whereas our task uses regression losses (EMD for GIAA, MSE for PIAA), whose magnitudes differ, we adjust a subset of coefficients when this scale gap prevents the method from operating as intended. \textbf{Third}, we follow the practical constraint of unsupervised adaptation by selecting all hyperparameters without using any target-domain validation set. Unless a parameter is specified separately for GIAA and PIAA, the same value is used for both tasks.

\paragraph{Adversarial methods (\textit{DANN}, \textit{CDAN}, \textit{ALDA}).}
We choose the annealing schedule under two diagnostic targets. For \textit{DANN} and \textit{CDAN}, we select the schedule length so that the domain-discriminator accuracy fluctuates around $0.5$, ensuring the features become domain-invariant without the adversarial game collapsing. For \textit{ALDA}, which adds a pseudo-label-based loss $\mathcal{L}_T$, we select the schedule length so that the source task loss ratio $\mathcal{L}_y / (\mathcal{L}_y + \lambda\mathcal{L}_T + \mathcal{L}_{\mathrm{Adv}} + \mathcal{L}_{\mathrm{Reg}})$ stays within $40$--$60\%$, avoiding over-reliance on the still-unreliable early-stage pseudo-labels.

These criteria yield a common schedule length of 100 epochs (with sigmoid sharpness $\gamma{=}10.0$). The remaining hyperparameters are fixed independently: the soft-ordinal width to $\sigma{=}1.0$ and, for \textit{ALDA}, the pseudo-label confidence threshold to $\delta{=}0.2$. The resulting warm-up lets the source representation stabilize before the adversarial loss takes full effect, avoiding the early-stage destabilization that the EMD/MSE task loss would otherwise induce.

\paragraph{Optimal-transport methods (\textit{DeepJDOT}, \textit{JUMBOT}).}
Both align the joint feature--label distribution via optimal transport. We choose the feature- and label-cost weights and, for \textit{JUMBOT}, the marginal-relaxation parameter under two diagnostic targets. For both methods, we tune the feature- and label-cost weights so that the source task loss remains dominant. For \textit{JUMBOT}, we set the marginal-relaxation parameter $\tau$ so that the total transport mass stabilizes within $0.7$--$0.9$ (active relaxation without collapse).

These criteria yield feature- and label-cost weights of $(\alpha, \lambda_t){=}(0.1, 0.1)$ for \textit{DeepJDOT} and $(\eta_1, \eta_2){=}(0.1, 0.1)$ for \textit{JUMBOT} on GIAA, and $(0.01, 1.0)$ and $(0.01, 0.5)$, respectively, on PIAA. For \textit{JUMBOT}, the relaxation parameter is $\tau{=}0.5$ on GIAA and $\tau{=}0.1$ on PIAA. The remaining hyperparameters are fixed independently: the transfer-loss scale to $\eta_3{=}1.0$ and the entropic regularizer to $\varepsilon{=}0.1$.

\paragraph{Feature-alignment methods (\textit{DeepCORAL}, \textit{RSD}, \textit{DARE-GRAM}).}
\textit{DeepCORAL} minimizes the distance between source and target covariance matrices with a fixed weight $\lambda_{\mathrm{coral}}$. \textit{RSD} and \textit{DARE-GRAM} extend feature alignment to regression and are therefore applied only to PIAA, where each is built on the \textit{DeepCORAL} model weights that performed best on GIAA in our benchmark; each introduces a primary alignment term plus a secondary stabilizer. We choose the alignment and stabilizer weights under two diagnostic targets. For \textit{DeepCORAL}, we set $\lambda_{\mathrm{coral}}$ so that the source task loss remains dominant; because the raw $\mathcal{L}_{\mathrm{coral}}$ is much smaller in scale than our regression loss, this requires a large weight. For \textit{RSD} and \textit{DARE-GRAM}, we tune the primary and secondary weights jointly so that the primary alignment term dominates: the \textit{RSD} term occupies $70$--$90\%$ of the alignment loss for \textit{RSD}, and the angle term $40$--$60\%$ for \textit{DARE-GRAM}.

These criteria yield $\lambda_{\mathrm{coral}}{=}10^{4}$ for \textit{DeepCORAL}; $(\beta, \gamma){=}(2.5\times10^{-3}, 2.5\times10^{-4})$ (the \textit{RSD} weight $\beta$ and the BMP weight $\gamma$) for \textit{RSD}; and $(\alpha_{\cos}, \gamma_{\mathrm{scale}}){=}(50, 10^{-4})$ for \textit{DARE-GRAM}. The remaining hyperparameters are fixed independently: the numerical-stabilization constant $\varepsilon{=}10^{-8}$ for \textit{RSD} and the truncation threshold $T{=}0.9$ for \textit{DARE-GRAM}.

\begin{table*}[t]
\centering
\caption{Comparison of backbone architectures on the GIAA task across three domains.}
\label{tab:comp_backbone}
\begin{tabular}{c|cc|cc|cc|cc}
\hline
\textbf{Backbone}
 & \multicolumn{2}{c|}{\textbf{Art}}
 & \multicolumn{2}{c|}{\textbf{Fashion}}
 & \multicolumn{2}{c|}{\textbf{Landscape}}
 & \multicolumn{2}{c}{\textbf{Avg.}} \\
\hline
 & EMD $\downarrow$ & SCC $\uparrow$
 & EMD $\downarrow$ & SCC $\uparrow$
 & EMD $\downarrow$ & SCC $\uparrow$
 & EMD $\downarrow$ & SCC $\uparrow$ \\
\hline
GPT 5.4
 & .333{\tiny$\pm$.004} & .673{\tiny$\pm$.022}
 & .390{\tiny$\pm$.010} & .508{\tiny$\pm$.032}
 & .339{\tiny$\pm$.007} & .535{\tiny$\pm$.040}
 & .354 & .572 \\
Claude Opus 4.6
 & .325{\tiny$\pm$.006} & .724{\tiny$\pm$.038}
 & .350{\tiny$\pm$.011} & .494{\tiny$\pm$.046}
 & .338{\tiny$\pm$.005} & .554{\tiny$\pm$.020}
 & .338 & .591 \\
Gemini 3.0 Flash
 & .312{\tiny$\pm$.008} & .693{\tiny$\pm$.038}
 & .327{\tiny$\pm$.010} & .505{\tiny$\pm$.035}
 & .382{\tiny$\pm$.009} & .496{\tiny$\pm$.020}
 & .340 & .565 \\
\hline
ResNet50
 & .293{\tiny$\pm$.002} & .728{\tiny$\pm$.027}
 & .319{\tiny$\pm$.012} & .449{\tiny$\pm$.045}
 & .302{\tiny$\pm$.011} & .667{\tiny$\pm$.025}
 & .305 & .615 \\
ViT-B/16
 & .298{\tiny$\pm$.005} & .709{\tiny$\pm$.026}
 & .319{\tiny$\pm$.012} & .442{\tiny$\pm$.029}
 & .301{\tiny$\pm$.013} & .658{\tiny$\pm$.026}
 & .306 & .603 \\
CLIP (ResNet50)
 & .\textbf{272}{\tiny$\pm$.007} & .789{\tiny$\pm$.030}
 & .302{\tiny$\pm$.008} & .551{\tiny$\pm$.027}
 & .279{\tiny$\pm$.007} & .749{\tiny$\pm$.010}
 & .284 & .696 \\
CLIP (ViT-B/16)
 & .272{\tiny$\pm$.005} & \textbf{.790}{\tiny$\pm$.027}
 & \textbf{.298}{\tiny$\pm$.007} & \textbf{.574}{\tiny$\pm$.037}
 & \textbf{.276}{\tiny$\pm$.006} & \textbf{.755}{\tiny$\pm$.016}
 & \textbf{.282} & \textbf{.706} \\
$\text{I3D}$
 & --- & ---
 & --- & ---
 & .300{\tiny$\pm$.007} & .668{\tiny$\pm$.012}
 & --- & --- \\
\hline
\end{tabular}
\vspace{-0.3em}
\begin{center}
\footnotesize
Values are mean\,$\pm$\,std over 5 folds. \textbf{Bold}: best per column. I3D is the only pretrained backbone that takes video as input.
\end{center}
\end{table*}

\section{Backbone Comparison on the GIAA Task}
\label{app:backbone}
In the main text, GIAA models are assessed only through their effect on the downstream PIAA task. Here, independently of that setting, we evaluate a range of backbones directly on the GIAA task itself, in order to understand which backbone best supports population-level aesthetic prediction.

\subsection{Setup}
We follow the same \textit{NIMA}-based GIAA training protocol as in the main experiments. The only departure from it lies in the data-splitting scheme. For this analysis we construct a data split tailored specifically to the GIAA task, so that backbone capacity for the population-level task is evaluated in isolation. The ratings of all $129$ annotators are aggregated into per-image $7$-point score histograms, and we perform $5$-fold cross-validation at the image level: in each fold, $20\%$ of the images are held out as the test set and the remaining $80\%$ form the training pool, $10\%$ of which is reserved as a validation set for early stopping and learning-rate scheduling. \textit{NIMA} is trained independently per domain.

\begin{figure*}[t]
    \centering
    \includegraphics[width=0.8\linewidth]{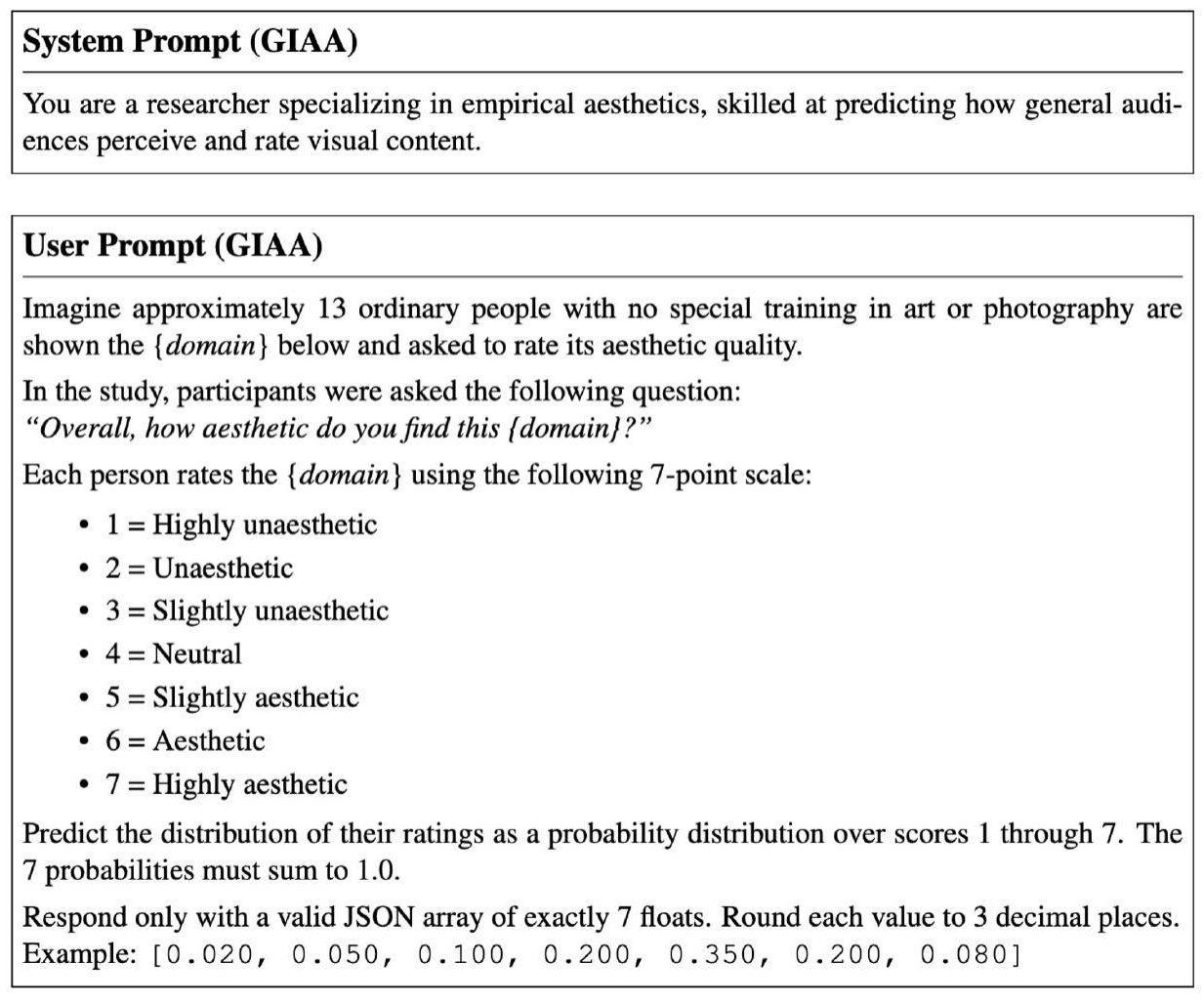}
    \vspace{0.2em}
\caption{Prompt templates used for zero-shot GIAA evaluation with proprietary LLMs.}
    \label{fig:prompt_GIAA}
\end{figure*}

We compare ImageNet-pretrained backbones (\textit{ResNet50}, \textit{ViT-B/16}), their \textit{CLIP}-pretrained counterparts with OpenAI weights, and the video backbone \textit{I3D}. \textit{I3D} is reported only for the Landscape domain, as only that domain contains video stimuli. For reference, we also report the three proprietary LLMs (\textit{GPT-5.4}, \textit{Claude Opus 4.6}, \textit{Gemini 3.0 Flash}) evaluated zero-shot using the prompt in Fig.~\ref{fig:prompt_GIAA}; these models are not trained on our data.

\subsection{Results and Discussion}
Table~\ref{tab:comp_backbone} reports the comparison. \textit{CLIP} (\textit{ViT-B/16}) attains the best average performance (EMD $.282$, SCC $.706$) and ranks first in most domains, with \textit{CLIP} (\textit{ResNet50}) a close second ($.284$\,/\,$.696$). The decisive factor is the \emph{pretraining} rather than the architecture: replacing ImageNet with \textit{CLIP} pretraining improves the average SCC from $.615$ to $.696$ for \textit{ResNet50} and from $.603$ to $.706$ for \textit{ViT-B/16}, whereas at a fixed pretraining the two architectures are nearly indistinguishable. We attribute this to the fact that \textit{CLIP} features, learned by aligning images with natural-language descriptions, encode semantic and stylistic cues that correlate with aggregated aesthetic judgment far better than purely supervised ImageNet features, consistent with recent findings on the usefulness of vision-language representations for aesthetics~\cite{ryu2026vl}.

The three proprietary LLMs, evaluated zero-shot, trail all trained backbones: the best of them (\textit{Claude Opus 4.6}, EMD $.338$\,/\,SCC $.591$) still falls short of the ImageNet-pretrained backbones. This gap highlights the value of labeled supervision: trained on annotated ratings, a supervised head calibrates to the dataset's score range and annotator tendencies---grounding that zero-shot prompting lacks, since it must infer the aggregate rating distribution from the prompt alone. Since this comparison uses a single zero-shot configuration without prompt optimization or few-shot exemplars, we read it as characterizing that setting and do not intend it to generalize to proprietary-LLM-based approaches more broadly.

\textit{I3D}, available only for landscape, performs on par with the ImageNet-pretrained \textit{ResNet50} there but well below the \textit{CLIP} variants. Because our video stimuli are one-minute clips recorded while walking and thus exhibit little temporal variation, the motion information \textit{I3D} captures adds little beyond a single frame, and single-frame semantic representations suffice for the population-level GIAA task. Across domains, fashion is consistently the hardest, exhibiting the lowest SCC for every backbone, which suggests that aesthetic ranking in that domain is less separable from global visual features alone; it may also reflect a relative scarcity of fashion-related content in the pretraining data, so that the learned representations capture fashion aesthetics less well. Taken together, these results motivate our choice of frozen \textit{CLIP} (\textit{ViT-B/16}) as the backbone for all GIAA and PIAA experiments in the main text.

\begin{figure*}[t]
    \centering
    \includegraphics[width=1\linewidth]{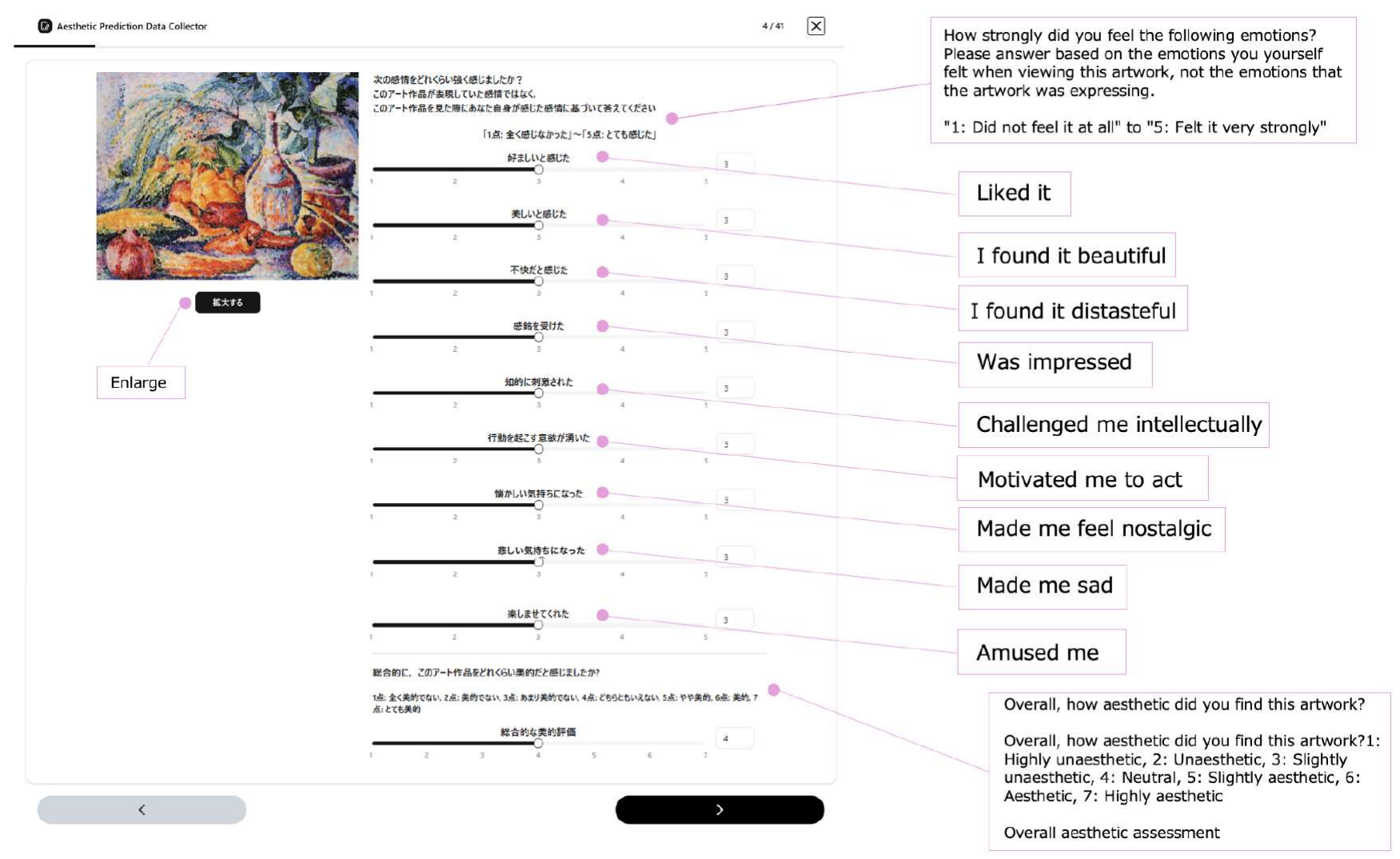}
    \vspace{0.2em}
\caption{%
    Screenshot of the web-based annotation interface, showing the nine aesthetic emotion items (5-point scale) and the overall aesthetic assessment (7-point scale) presented for each stimulus.
}
    \label{fig:annotation_tool}
\end{figure*}

\begin{figure*}[t]
    \centering
    \includegraphics[width=0.9\linewidth]{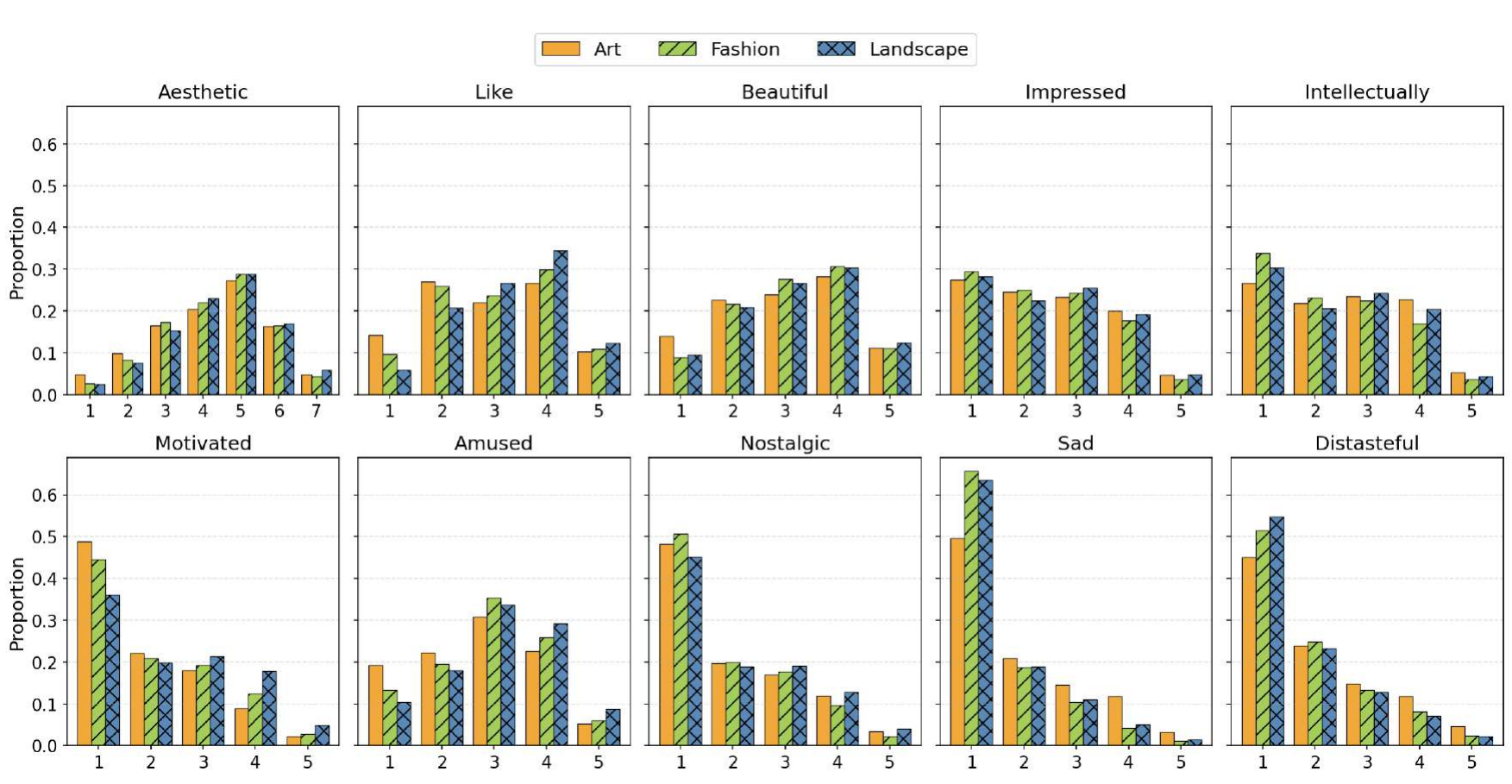}
    \vspace{0.2em}
\caption{%
    Per-item rating distributions for the three aesthetic-evaluation items (Aesthetic, Like, Beautiful) and seven aesthetic-emotion items on a 1–7 Likert scale, shown separately for the three domains (art, fashion, landscape). Bars give the within-domain proportion of responses at each scale point and share a common y-axis across panels.
}
    \label{fig:rating_histograms}
\end{figure*}

\begin{table}[t]
\centering
\caption{Descriptive statistics of all annotation items across three domains.}
\label{tab:item_stats}
\vspace{0.5em}
\footnotesize
\begin{threeparttable}
\begin{tabular}{lrrrrrr}
\toprule
& \multicolumn{2}{c}{Art} & \multicolumn{2}{c}{Fashion} & \multicolumn{2}{c}{Landscape} \\
\cmidrule(lr){2-3}\cmidrule(lr){4-5}\cmidrule(lr){6-7}
Item & Mean & SD & Mean & SD & Mean & SD \\
\midrule
Like                 & 2.92 & 1.23 & 3.06 & 1.17 & 3.26 & 1.10 \\
Beautiful            & 3.00 & 1.23 & 3.13 & 1.14 & 3.15 & 1.17 \\
Impressed            & 2.50 & 1.22 & 2.41 & 1.18 & 2.50 & 1.22 \\
Intellectually       & 2.58 & 1.24 & 2.33 & 1.20 & 2.48 & 1.23 \\
Motivated            & 1.93 & 1.10 & 2.08 & 1.18 & 2.36 & 1.27 \\
Amused               & 2.72 & 1.16 & 2.92 & 1.11 & 3.08 & 1.11 \\
Nostalgic            & 2.03 & 1.20 & 1.93 & 1.12 & 2.12 & 1.23 \\
Sad                  & 1.98 & 1.19 & 1.56 & 0.91 & 1.62 & 0.96 \\
Distasteful          & 2.07 & 1.21 & 1.85 & 1.08 & 1.78 & 1.05 \\
\bottomrule
\end{tabular}
\begin{tablenotes}
\footnotesize
\item Values are reported on the display scale. The overall aesthetic assessment uses a 7-point scale; all other items use a 5-point scale.
\end{tablenotes}
\end{threeparttable}
\end{table}

\begin{figure*}[t]
    \centering
    \includegraphics[width=0.8\linewidth]{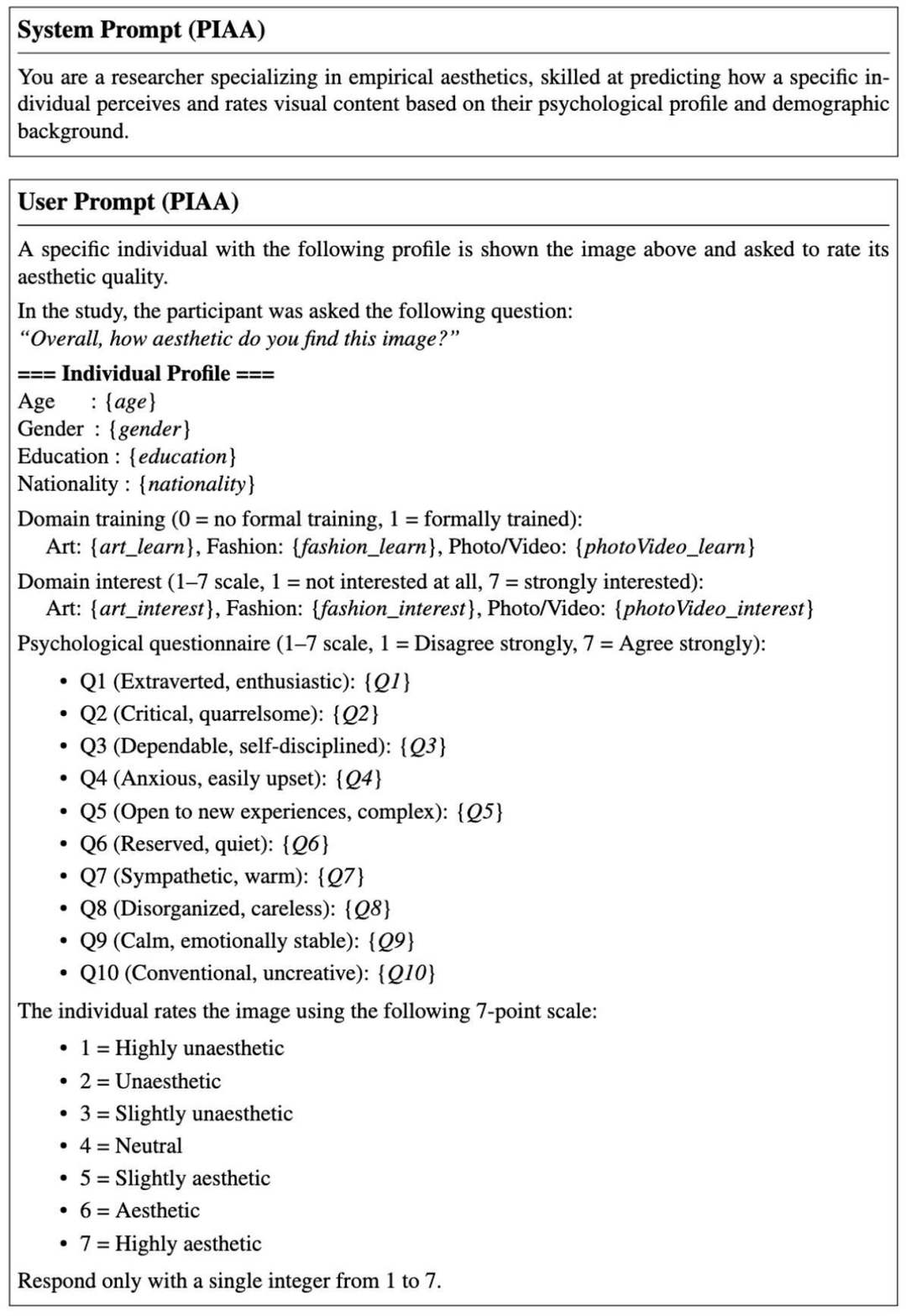}
    \vspace{0.2em}
\caption{Prompt templates used for zero-shot PIAA evaluation with proprietary LLMs.}
    \label{fig:prompt_PIAA}
\end{figure*}

\end{document}